\documentclass[ejs,preprint]{imsart}

\RequirePackage[OT1]{fontenc}
\RequirePackage{amsthm,amsmath}
\usepackage{amssymb,amsfonts, amsbsy, epsfig, subfig, graphicx,rotating}
\usepackage[authoryear]{natbib}
\RequirePackage[colorlinks,citecolor=blue,urlcolor=blue]{hyperref}

\pubyear{2016}
\volume{10}
\issue{}
\firstpage{1493}
\lastpage{1525}
\doi{10.1214/16-EJS1147}

\startlocaldefs

\newcommand{\bfsym}[1]{\ensuremath{\boldsymbol{#1}}}

\def\bxi       {\bfsym {\xi}}

\def\beps      {\bfsym \varepsilon}

\newcommand{\noi}{\noindent}

\newcommand{\la}{\lambda}

\newcommand{\beq}{\begin{eqnarray*}}
\newcommand{\eeq}{\end{eqnarray*}}
\newcommand{\beqn}{\begin{eqnarray}}
\newcommand{\eeqn}{\end{eqnarray}}

\newcommand{\var}{\mbox{Var}}
\newcommand{\bi}{\begin{itemize}}
\newcommand{\ei}{\end{itemize}}

\newcommand{\be}{\begin{equation}}
\newcommand{\ee}{\end{equation}}
\newcommand{\nn}{\nonumber}

\newcommand{\lbl}{\label}

\newcommand{\eq}[1]{$(\ref{#1})$}

\newcommand{\ignore}[1]{}{}
\newcommand{\f}{\frac}

\newcommand{\sn}{\sum_{i=1}^n}

\renewcommand{\P}{\mathbb{P}}

\newcommand{\de}{\delta}
\newcommand{\De}{\Delta}
\newcommand{\ga}{\gamma}

\newcommand{\s}{\sqrt}

\newcommand{\e}{\mathbb{E}}
\newcommand{\p}{\mathbb{P}}

\newcommand{\br}{\mathbb{R}}
\newcommand{\argmin}{\mathop{\rm arg\min}}

\numberwithin{equation}{section}
\theoremstyle{plain}
\newtheorem{lemma}{Lemma}[section]

\newtheorem{theorem}{Theorem}[section]

\newtheorem{corollary}{Corollary}[section]
\newtheorem{remark}{Remark}[section]
\newtheorem{definition}{Definition}[section]

\numberwithin{equation}{section}
\theoremstyle{plain}

\endlocaldefs

\begin{document}

\begin{frontmatter}
\title{Matrix completion via max-norm constrained optimization}
\runtitle{Matrix completion via max-norm regularizaion}

\begin{aug}

\author{\fnms{T. Tony} \snm{Cai}\thanksref{t1}\ead[label=e1]{tcai@wharton.upenn.edu}}

\address{Department of Statistics, The Wharton School \\
 University of Pennsylvania, Philadelphia, PA 19104, USA \\
\printead{e1} }

\medskip
\and

\author{\fnms{Wen-Xin} \snm{Zhou}\thanksref{t2}
\ead[label=e3]{wenxinz@princeton.edu}
\ead[label=u1,url]{www.foo.com}}

\address{Department of Operations Research and Financial Engineering\\
Princeton University, Princeton, NJ 08544, USA\\
\printead{e3}}

\thankstext{t1}{Supported in part by NSF Grants DMS-1208982 and DMS-1403708, and NIH Grant R01 CA127334.}
\thankstext{t2}{Supported in part by NIH Grant R01-GM100474-5.}

\runauthor{T. T. Cai and W.-X. Zhou}

\affiliation{University of Pennsylvania and Princeton University}

\end{aug}

\begin{abstract}
Matrix completion has been well studied under  the uniform sampling model and the trace-norm regularized methods perform well both  theoretically and numerically in such a setting. However,  the uniform sampling model is unrealistic for a range of applications and the standard trace-norm relaxation can behave very poorly when the underlying sampling scheme is non-uniform.

In this paper we propose and analyze a max-norm constrained empirical risk minimization method for noisy matrix completion under a general sampling model. The optimal rate of convergence is established under the Frobenius norm loss in the context of approximately low-rank matrix reconstruction. It is shown that the max-norm constrained method is minimax rate-optimal and yields a unified and robust approximate recovery guarantee, with respect to the sampling distributions. The computational effectiveness of this method is also discussed, based on first-order algorithms for solving convex optimizations involving max-norm regularization.
\end{abstract}


\tableofcontents
\end{frontmatter}

\section{Introduction}

The problem of recovering a low-rank matrix from a subset of its entries, also known as \textit{matrix completion}, has been an active topic of recent research with a range of applications including collaborative filtering (the Netflix problem) \citep{GNOT92}, multi-task learning \citep{AEP08}, system identification \citep{LV09}, and sensor localization \citep{SC10, CP10}, among many others. We refer to \cite{CP10} for detailed discussions of the aforementioned applications. 
Another noteworthy example is the structure-from-motion problem in computer vision \citep{TK92, CS04}.  Let $f$ and $d$ be the number of frames and feature points, respectively. The data are stacked into a low-rank matrix of trajectories, say $M \in \br^{2f \times d}$, such that every element of $M$ corresponds to an image coordinate from a feature point of a rigid moving object at a given frame. Due to objects occlusions, errors on the tracking or variable out of range (i.e. images beyond the camera field of view), missing data are inevitable in real-life applications and are represented as empty entries in the matrix. Therefore, accurate and effective matrix completion methods, which fill in missing entries by suitable estimates, are required.

Because a direct search for the lowest-rank matrix satisfying the equality constraints is NP-hard, most previous work on matrix completion has focused on using the trace-norm, which is defined to be the sum of the singular values of the matrix, as a convex relaxation for the rank. This can be viewed as an analogue to relaxing the {\it sparsity} of a vector to its $\ell_1$-norm, which has been shown to be effective both empirically and theoretically in compressed sensing.  Several recent papers proved in different settings that a generic $d\times d$ rank-$r$ matrix can be exactly and efficiently recovered from $O\{ r d \, \mbox{poly}( \log d  )\}$ randomly chosen entries \citep{CR09, CT10, G11, R11}. These results thus provide theoretical guarantees for the constrained trace-norm minimization method. In the case of recovering approximately low-rank matrices based on noisy observations, different types of trace-norm based estimators, which are akin to the Lasso and Dantzig selector used in sparse signal recovery, were proposed and well-studied. See, for example, \cite{CP10}, \cite{KAO10}, \cite{RT11}, \cite{KLT2011}, \cite{NW2012}, \cite{K12} and \cite{Klo11, Klo14}, among others.
 
It is, however, unclear that whether the trace-norm is the best convex relaxation for the rank, especially when the underlying sampling scheme is non-uniform, and more importantly, is unknown. A matrix $M \in \br^{d_1 \times d_2}$ can be viewed as an operator mapping from $\br^{d_2}$ to $\br^{d_1}$, its rank can be alternatively expressed as the smallest integer $k$ such that the matrix $M$ can be decomposed as $M=U V^{\intercal}$ for some $U \in \br^{d_1\times k}$ and $V \in \br^{d_2 \times k}$. In view of the matrix factorization $M=UV^{\intercal}$, by enforcing $U$ and $V$ to have a small number of columns we obtain a low-rank $M$. The number of columns of $U$ and $V$ can be relaxed in a different way from the usual trace-norm by the so-called {\it max-norm} \citep{LMSS04}, defined by
\be 
	\| M \|_{\max} = \min_{M=UV^{\intercal}}   \| U \|_{2, \infty} \| V \|_{2,\infty} , 	 \label{eq1.1}
\ee
where the infimum is carried out over all factorizations $M=UV^{\intercal}$ with $\| U \|_{2, \infty}$ denoting the operator norm of $U: \ell_2^k \mapsto \ell^{d_1}_{\infty}$ and $ \| V \|_{2,\infty} $ the operator norm of $V:\ell_2^k \mapsto \ell_{\infty}^{d_2}$ (or, equivalently, $V^{\intercal}: \ell^{d_2}_1 \mapsto \ell_2^k$) and $k=1, \ldots, \min(d_1, d_2)$. Note that $\|U \|_{2,\infty}$ is also the maximum $\ell_2$ row norm of $U$. Since $\ell_2$ is a Hilbert space, the factorization constant $\| \cdot \|_{\max}$ indeed defines a norm on the space of operators between $\ell^{d_2}_1$ and $\ell_{\infty}^{d_1}$.
 
The max-norm was recently proposed as an alternative convex surrogate to the rank of the matrix. For collaborative filtering problems, the max-norm has been shown to be empirically superior to the trace-norm \cite{Sre04}. \cite{FS2011} used the max-norm for matrix completion under the uniform sampling distribution. Their results are direct consequences of a recent bound on the excess risk for a smooth loss function, such as the quadratic loss, with a bounded second derivative \citep{SST10}. Further, a max-norm constrained maximum likelihood method was considered by \cite{CZ2013} for one-bit matrix completion, where instead of observing real-valued entries of an unknown matrix one is only able to see binary outputs, i.e. yes/no, true/false, agree/disagree \citep{DP}. Theoretical guarantees are obtained in general non-uniform sampling models, and numerical studies show that the max-norm based approach is comparable to and sometimes slightly outperform the corresponding trace-norm method.

Matrix completion has been well analyzed in the uniform sampling model, where observed entries are assumed to be sampled randomly and uniformly. In such a setting, the trace-norm regularized approach has been shown to have good  theoretical and numerical  performance. However, in some applications such as collaborative filtering, the uniform sampling model is unrealistic. For example, in the Netflix problem, the uniform sampling model is equivalent to assuming all users are equally likely to rate each movie and all movies are equally likely to be rated by any user. From a practical point of view, invariably some users are more active than others and some movies are more popular and thus rated more frequently. Hence, the sampling distribution is in fact non-uniform in the real world. In such a setting, \cite{SS10} showed that the standard trace-norm relaxation can sometimes behave poorly, and suggested a weighted trace-norm penalty, which incorporates the knowledge of true sampling distribution in its construction. Since the true sampling distribution is  most likely unknown and can only be estimated based on the locations of those entries that are revealed in the sample, a practically available method relies on the empirically-weighted trace-norm \citep{FSSS}. It is also worth noticing that, when the sampling probabilities are bounded from below and above, the trace-norm penalized estimator is minimax optimal up to a logarithmic factor \citep{Klo14}. We refer to \cite{F2015b} for further numerical evaluations of the trace-norm regularized method under various non-uniform sampling schemes.

In this paper, we employ the max-norm as a convex relaxation for the rank to study matrix completion based on noisy observations in a general, unspecified sampling model. The rate of convergence for the max-norm constrained least squares estimator is obtained.
Information-theoretical methods are used to establish a matching minimax lower bound in the general non-uniform sampling model.
Together, the minimax upper and lower bounds yield the optimal rate of convergence for the Frobenius norm loss. It is shown that the max-norm regularized approach indeed provides a unified and robust approximate recovery guarantee with respect to sampling schemes. In the uniform sampling model as a special case, our results also show that the extra logarithmic factors appeared in the error rates obtained by \cite{SST10} and \cite{FS2011} could be avoided after a careful analysis to match the minimax lower bound with the upper bound (see Theorems \ref{Thm3.1} and  \ref{Thm3.3} and the discussions in Section \ref{main.sec}). 

The max-norm constrained minimization problem is a convex program. To solve general convex programs that involve either a max-norm constraint or a max-norm penalization, a first-order algorithm was proposed by \cite{Lee2010}, which is computationally effective and outperforms the semi-definite programming (SDP) method of \cite{Sre04}. In principle, the method of \cite{Lee2010} is based on nonconvex relaxations. Therefore, their algorithm is only guaranteed to find a stationary point, and statistical properties of such solutions are difficult to analyze. Recently, \cite{F2015b} proposed a scalable algorithm based on the alternating direction of multipliers method to efficiently solve the max-norm constrained optimization problem with guaranteed rate of convergence to the global optimum. In summary, the max-norm constrained empirical risk minimization problem can indeed be implemented in polynomial time as a function of the sample size and matrix dimensions.

The remainder of the paper is organized as follows. After introducing basic notation and definitions, Section \ref{notation.sec} collects a few useful results on the max-norm, trace-norm and Rademacher complexity that will be needed in the rest of the paper. Section \ref{main.sec} introduces the model and the estimation procedure and then investigates the theoretical properties of the estimator. Both minimax upper and lower bounds are given. The results show that the max-norm constrained minimization method achieves the optimal rate of convergence over the parameter space. Comparison with past work is also given. Computation and implementation issues are discussed in Section \ref{implementation.sec}. A brief discussion is given in Section~\ref{disc.sec}, and the proofs of the main results and key technical lemmas are placed in Section \ref{proof.sec}.

\section{Notations and Preliminaries}
\label{notation.sec}

In this section, we begin with some notation that will be used throughout the paper, and then collect some known results on the max-norm, trace-norm and Rademacher complexity that will be applied repeatedly later.  
 
For any positive integer $d$, we use $[d]$ to denote the collection of integers $\{ 1, 2, \ldots, d\}$.  For any set $S$, denote by $S^c$ its complement, and $|S|$ its cardinality. For a vector $u\in \br^d$ and $1\leq p < \infty$, define its $\ell_p$-norm by $\| u \|_p=\big( \sum_{i=1}^d |u_i|^p \big)^{1/p}$. In particular, $\| u \|_{\infty}=\max_{i=1, \ldots, d}| u_i|$ is the $\ell_{\infty}$-norm. For any $d_1 \times d_2$ matrix $M =(M_{k \ell })_{1\leq k\leq d_1, 1\leq \ell \leq d_2}$, let $\| M \|_F = \s{\sum_{k=1}^{d_1} \sum_{ \ell =1}^{d_2} M_{k \ell }^2}$ be the Frobenius norm and let $\| M\|_{\infty}=\max_{(k,\ell)\in [d_1]\times [d_2]}|M_{k \ell }|$ denote the elementwise $\ell_{\infty}$-norm. Given two norms $\ell_p$ and $\ell_q$ on $\br^{d_1}$ and $\br^{d_2}$ respectively, the corresponding operator norm $\| \cdot \|_{p,q}$ of a matrix $M\in \br^{d_1 \times d_2}$  is defined by $\| M \|_{p,q}=\sup_{\| u \|_p=1} \| M u\|_q$.  It is easy to verify that $\| M\|_{p,q}=\| M^{\intercal} \|_{q^*,p^*}$, where $(p,p^*)$ and $(q, q^*)$ are conjugate pairs; that is, $1/p+1/p^* = 1/q+1/q^*=1$. In particular, $\| M \| = \| M \|_{2,2}$ is the spectral norm; $\| M \|_{2,\infty}=\max_{k=1, \ldots, d_1} \s{\sum_{\ell =1}^{d_2} M_{k \ell }^2}$ is also known as the maximum row norm of $M$. Moreover, for two real numbers $a$ and $b$, we write for ease of presentation that $a \vee b=\max(a, b)$ and $a \wedge b=\min(a, b)$.

\subsection{Max-norm and trace-norm} 
\label{pre-1}

For a matrix $M \in \br^{d_1 \times d_2}$, the trace-norm (also known as the Schatten $1$-norm)  $\| M \|_1$ is defined as the sum of all singular values of $M$, or equivalently,
\begin{align*}
& \| M \|_1 \\
&  = \inf\left\{ \sum_{j=1}^{d_1 \wedge d_2}  |\sigma_j | :  M = \sum_{j=1}^{d_1 \wedge d_2} \sigma_j u_j v_j^{\intercal}, \, u_j \in \br^{d_1}, v_j \in \br^{d_2}, \,  \| u_j \|_2 = \| v_j \|_2 =1 \right\}.
\end{align*}
In other words, the trace-norm promotes low-rank decompositions with factors in $\ell_2$. Similarly, using Grothendiek's inequality \citep{J87},  the max-norm defined in \eqref{eq1.1} has the following analogous representation in terms of factors in $\ell_{\infty}$:
\begin{align*}
	\| M \|_{\max}   \approx  \inf\left\{  \sum_{j=1}^{d_1 \wedge d_2} |\sigma_j | :  M = \sum_{j=1}^{d_1 \wedge d_2} \sigma_j u_j v_j^{\intercal},   \,   \| u_j \|_{\infty} = \| v_j \|_{\infty} =1 \right\}.
\end{align*}
The factor of equivalence is the Grothendieck's constant $K_G \in (1.67, 1.79)$. Based on these properties, the max-norm regularization is expected to be more effective when dealing with uniformly bounded data \citep{Lee2010}.

Of the same spirit as the definition of the max-norm in \eqref{eq1.1}, the trace-norm has the following equivalent characterization in terms of matrix factorizations:
$$
	\| M \|_1 = \min_{U, V: M = U V^{\intercal} }   \| U \|_F \| V \|_F  = \f{1}{2} \min_{U, V: M = U V^{\intercal}} \left( \| U \|_F^2 + \| V \|_F^2 \right).
$$
See, for example, \cite{SS2005}. It is easy to see that
\be
	\f{1}{\s{d_1 d_2}} \| M \|_1  \leq \| M \|_{\max},  \label{eq2.1}
\ee
which in turn implies that any low max-norm approximation is also a low trace-norm approximation. As pointed out by \cite{SS2005}, there can be a large gap between $\f{1}{\s{d_1 d_2}}\| \cdot \|_1$ and $\| \cdot \|_{\max}$. The following relation between the trace-norm and Frobenius norm is well-known: $\|M \|_F \leq \| M \|_1 \leq \s{\mbox{rank}(M)} \cdot \| M \|_F$. An analogous bound holds for the max-norm, in connection with the element-wise $\ell_{\infty}$-norm \citep{LMSS04}:
\be 
	\| M \|_{\infty } \leq \| M \|_{\max} \leq \s{\mbox{rank}(M)} \cdot \| M \|_{1,\infty} \leq \s{\mbox{rank}(M)} \cdot \| M \|_{\infty}.  \label{eq2.2}
\ee
For any $R>0$, let
\begin{align}
	\mathbb{B}_{\max}(R)    = \big\{ M \in \br^{d_1 \times d_2}: \| M \|_{\max} \leq R  \big\}  \nn \\
 \mbox{ and }  \ \ \mathbb{B}_{{\rm tr}}(R)   = \big\{ M \in \br^{d_1 \times d_2}: \| M \|_1 \leq R  \big\} \nn
\end{align}
be the max-norm and trace-norm ball with radius $R$, respectively. It is now well-known \citep{SS2005} that $\mathbb{B}_{\max}(1)$ can be bounded, from both below and above, by the convex hull of rank-one sign matrices $\mathcal{M}_{\pm}= \{ M \in \{\pm 1 \}^{d_1 \times d_2}: \mbox{rank}(M)=1\}$, i.e.
\be
	 \mbox{conv} \mathcal{M}_{\pm} \subseteq \mathbb{B}_{\max}(1)  \subseteq  K_G \cdot \mbox{conv} \mathcal{M}_{\pm}  \label{eq2.3}
\ee
with $ K_G \in (1.67, 1.79)$ denoting the Grothendieck's constant. Moreover, $\mathcal{M}_{\pm}$ is a finite class with cardinality $|\mathcal{M}_{\pm}|=2^{d-1}$, where $d=d_1+d_2$.

\medskip
\subsection{Rademacher complexity}

A technical tool used in our analysis involves data-dependent estimates of the Rademacher and Gaussian complexities of a function class. We refer to \cite{BM2001} and \cite{SS2005} for a detailed introduction of these concepts.

\begin{definition} \label{Def2.1}
For a class $\mathcal{F}$ of functions mapping from $\mathcal{X}$ to $\mathbb{R}$, its empirical Rademacher complexity over a specific sample $S=(x_1, x_2, \ldots , x_n ) \subseteq \mathcal{X}$ is given by
\be
	\widehat{R}_{S}(\mathcal{F}) = \f{2}{|S|}  \e_{\beps} \left\{  \sup_{f\in \mathcal{F}} \left| \sn \varepsilon_i f(x_i) \right| \right\} ,  \nn
\ee
where $\beps =(\varepsilon_1, \varepsilon_2, \ldots , \varepsilon_n )^{\intercal}$ is a Rademacher sequence. The Rademacher complexity with respect to a distribution $\mathcal{P}$ is the expectation, over an independent and identically distributed (i.i.d.) sample of $|S|$ points drawn  from $\mathcal{P}$, denoted by
$$
	R_{|S|} (\mathcal{F}) =	\e_{S \sim \mathcal{P}} \{ \widehat{R}_{S}(\mathcal{F})\}.
$$
Replacing $\varepsilon_1, \ldots, \varepsilon_n$ with independent standard normal  variables $g_1, \ldots, g_n$ leads to the definition of (empirical) Gaussian complexity.
\end{definition}

Considering a matrix as a function from the index pairs to the entry values, \cite{SS2005} obtained upper bounds on the Rademacher complexity of the unit balls under both the trace-norm and the max-norm. Specifically, for any $d_1, d_2>2$ and any sample of size $2<|S|<d_1 d_2$, the empirical Rademacher complexity of the max-norm unit ball is bounded by
\be 
	\widehat{R}_S \left( \mathbb{B}_{\max}(1) \right) \leq 12  \s{ \f{d_1 + d_2}{|S|}}. \label{eq2.4}
\ee

\section{Max-Norm Constrained Empirical Risk Minimization}
\label{main.sec}
\setcounter{equation}{0}


\subsection{The statistical model}

We now consider matrix completion under a general random sampling model. Let $M^* \in \br^{d_1 \times d_2}$ be the unknown target matrix. Suppose that a random sample 
$$
	S=\{(i_1, j_1), (i_2, j_2), \ldots, (i_n,j_n)\} \subseteq \left(  [d_1] \times [d_2] \right)^n
$$ 
of the index set is drawn independently according to a general sampling distribution $\Pi=\{\pi_{k \ell}\}_{1\leq k\leq d_1, 1\leq \ell  \leq d_2}$ on $[d_1] \times [d_2]$, with replacement; that is, $\p \{ (i_t,j_t)=(k, \ell ) \}=\pi_{k \ell }$ for all $t=1,\ldots, n$ and $(k, \ell ) \in [d_1] \times [d_2]$. Given a random index subset $S=\{ (i_1, j_1), \ldots, (i_n, j_n) \}$ of size $n$, we observe noisy entries $\{Y_{i_t  j_t}\}_{t=1}^n$ indexed by $S$, i.e.
\be
	Y_{i_t  j_t} = M^*_{i_t  j_t} + \sigma \xi_t, \quad   t=1, \ldots, n,   \label{mc-md}
\ee
for some $\sigma>0$. The noise variables $\xi_t$ are independent with zero mean and unit variance. By expressing the model as in \eqref{mc-md}, it is implicitly assumed that the noise on the entry is drawn independently each time.

Instead of assuming the uniform sampling distribution, we consider  a general sampling distribution $\Pi$ here.
Since $\sum_{k=1}^{d_1}\sum_{ \ell =1}^{d_2} \pi_{k \ell} =1$, we have $\max_{k,\ell }  \pi_{k \ell }  \geq ( d_1 d_2)^{-1}$. Motivated by some applications, to ensure that each entry is observed with a positive probability, it is sometimes natural to assume that there exists a positive constant $\nu \geq 1$ such that
\be 
	\pi_{k  \ell } \geq \f{1}{\nu \, d_1 d_2}   \label{ass1}
\ee
holds for all $(k ,\ell ) \in [d_1]\times [d_2]$. We write hereafter $d=d_1 + d_2$ for brevity. Clearly, $\max(d_1, d_2) \leq d\leq 2\max(d_1, d_2)$.

The rescaled Frobenius norm $(d_1 d_2)^{-1} \| \cdot \|_F^2$ is typically used in the literature as a natural measure of the estimation accuracy. Now that the sampling distribution $\Pi$ is arbitrary, we use instead the weighted Frobenius norm with respect to $\Pi$ to measure the estimation error. For any $ A=(A_{k  \ell}) \in \br^{d_1\times d_2}$, define
\be  \label{weighted.Fro.norm}
	 \| A \|_{\Pi}^2 = \e_{(i,j) \sim \Pi} A^2_{ij}  = \sum_{k=1}^{d_1}\sum_{ \ell =1}^{d_2} \pi_{k \ell }  A_{k  \ell}^2. 
\ee
When $\Pi$ corresponds to the uniform distribution, $\| A \|_{\Pi}=(d_1 d_2)^{-1/2} \| A \|_F$.

The preceding work on matrix completion has mainly focused on the case of exact low-rank matrices. Here we allow a relaxation of this assumption and consider the more general setting of approximately low-rank matrices. Specifically, we consider recovery of matrices with $\ell_{\infty}$-norm and max-norm constraints defined by
\be 
	\mathcal{K}(\alpha , R) :=  \big\{ M\in \br^{d_1 \times d_2} : \| M \|_{\infty} \leq \alpha, \| M \|_{\max} \leq R  \big\}. \label{mc-set}
\ee
Here both $\alpha$ and $R$ are free parameters to be determined. If the matrix $M^*$ is of  rank at most $r$ and $\| M^* \|_{\infty} \leq \alpha$, then by \eq{eq2.2} we have $M^* \in \mathbb{B}_{\max}(\alpha\s{r})$ and hence $M^* \in \mathcal{K}(\alpha, \alpha \s{r})$.

\subsection{Max-norm constrained least squares estimator}

Given a collection of observations $Y_S=\{ Y_{i_t j_t} \}_{t=1}^n $ from the observation model \eq{mc-md}, we estimate the unknown $M^* \in \mathcal{K}(\alpha, R)$ for some $\alpha, R >0$ by the minimizer of the empirical risk with respect the quadratic loss function
\beq 
	\widehat{\mathcal{L}}_n(M; Y) = \f{1}{n} \sum_{t=1}^n (Y_{i_t  j_t} - M_{i_t  j_t} )^2.
\eeq
That is, 
\be 
	\widehat{M}_{\max} := \argmin_{M \in \mathcal{K}(\alpha, R)}  \widehat{\mathcal{L}}_n(M; Y). 
	\label{max-est}
\ee
The minimization procedure requires that all the entries of $M^*$ are bounded in magnitude by a prespecified constant $\alpha$.
This condition enforces that $M^*$ should not be too ``spiky'', and a too large bound may jeopardize exactness of the estimation. See, for example, \cite{KLT2011}, \cite{NW2012} and \cite{Klo14}. On the other hand, as argued in \cite{Lee2010}, the max-norm regularization is expected to be more effective particularly for uniformly bounded data, which is our main motivation for using the max-norm constrained estimator.

Although the max-norm constrained minimization problem \eq{max-est} is a convex program, 
fast and efficient algorithms for solving large-scale optimization problems that incorporate the max-norm have only been developed recently in \cite{Lee2010} and \cite{F2015b}. We will show in Section \ref{implementation.sec} that the convex optimization problem \eq{max-est} can be implemented in  polynomial time as a function of the sample size $n$ and dimensions $d_1$ and $d_2$.

\subsection{Upper bounds}

In this section, we state our main results regarding the recovery of an approximately low-rank (low-max-norm) matrix $M^*$ using max-norm constrained empirical risk minimization.

\begin{theorem} \label{Thm3.1}
Suppose that the noise sequence $\{\xi_t\}_{t=1}^n$ are independent sub-exponential random variables; that is, there is a constant $K>0$ such that
\be
	\max_{1\leq t\leq n} \e \{ \exp( |\xi_t|/K) \}  \leq e.   \label{sub-exp}
\ee
The parameters $\alpha, R>0$ are such that $M^*\in \mathcal{K}(\alpha, R)$. Then, for a sample size $n$ satisfying $d \leq  n \leq d_1 d_2$, 
\be 
	  \| \widehat{M}_{\max}-M^* \|_{\Pi}^2    
	\leq C   (\alpha \vee  K \sigma ) R \s{\f{d}{n}} ,  \label{mc-ubd-gen}
\ee 
with probability greater than $1-2e^{-d}$, where $C>0$ is an absolute constant. If, in addition, assumption \eq{ass1} is satisfied, then for a sample size $n$ with $d \leq  n \leq d_1 d_2$, 
\be 
	\f{1}{d_1 d_2} \| \widehat{M}_{\max} -M^* \|_F^2 \leq C \,\nu (\alpha \vee K\sigma )  R \s{\f{d}{n}} \label{mc-ubd2}
\ee
holds with probability at least $1-2e^{-d}$.
\end{theorem}

\medskip
\begin{remark}{\rm
\ \ 
\begin{itemize}
\item[(1)] It is worth noticing that the general result on approximate reconstruction guarantee \eqref{mc-ubd-gen} holds without any prior information on the sampling distribution $\Pi$, in particular the lower bound assumption \eqref{ass1}. In fact, it is reflected in the result that for every location index $(k,\ell)$, the smaller the sampling probability $\pi_{k \ell}$ is, the more difficult it will be to recovery the entry at this location.

\item[(2)] The upper bounds given in Theorem \ref{Thm3.1} hold with high probability. The rate of convergence under expectation can be obtained as a direct consequence. More specifically,  for a sample size $n$ with  $d\leq n\leq d_1 d_2$, we have
\be
	\sup_{M^* \in \mathcal{K}(\alpha, R)}	\f{1}{d_1 d_2} \e \| \widehat{M}_{\max} -M^* \|_F^2  \leq  C \, \nu (\alpha \vee \sigma )  R \s{\f{d}{n}}. \label{mc-ubd3}
\ee
\end{itemize}
}
\end{remark}

In view of the upper bound in \eqref{mc-ubd}, when the noise level $\sigma$ is comparable to or dominated by $\alpha$, the rate is of order $\alpha  R\, (\frac{d}{n})^{1/2}$. To fully understand how the random noise affects the estimation accuracy particularly when $\sigma$ is much smaller than $\alpha$, we provide a complementary result in Theorem~\ref{Thm3.2} which generalizes Theorem~9 in \cite{FS2011} to the general non-uniform sampling model.

\begin{theorem} \label{Thm3.2}
Assume that the conditions of Theorem~\ref{Thm3.1} are satisfied and $\sigma \leq \alpha$. Then,
\begin{align}
	& \| \widehat{M}_{\max} - M^* \|_{\Pi}^2 \nn \\
	&  \leq C \left\{  \sigma \sqrt{  (\log n)^3  \frac{  R^2 d   }{n} + (\log n)^{3/2} \frac{\alpha^2}{n} } +   (\log n)^3  \frac{R^2 d}{n} + (\log n)^{3/2} \frac{\alpha^2}{n}   \right\}  \label{error.bound.2}
\end{align}
holds with probability at least $1-2 n^{-1}$ over a random sample of size $n$ satisfying $d\leq n \leq d_1 d_2$, where $C > 1$ is a constant.
\end{theorem}

An interesting consequence of Theorem~\ref{Thm3.2} is that, in the noiseless case where $\sigma=0$ and a random subset of the entries of $M^*$ are perfectly observed, then for any prespecified tolerance level $\epsilon>0$, the target matrix $M^*$ can be approximately recovered in the sense that $\| \widehat{M}_{\max} - M^* \|_{\Pi}^2 \leq \epsilon$ whenever the sample size $n  \gtrsim \max\big\{  \frac{R^2 d}{\epsilon} (\log n)^3 , \frac{\alpha^2}{\epsilon} ( \log n )^{3/2} \big\}$.

\subsection{Information-theoretic lower bounds}

Theorem~\ref{Thm3.1} gives the rate of convergence for the max-norm constrained least squares estimator $\widehat{M}_{\max}$.  In this section we shall use information-theoretical methods to establish a minimax lower bound  for \textit{non-uniform sampling at random} matrix completion on the max-norm ball. The minimax lower bound matches the rate of convergence given in \eq{mc-ubd2} when the sampling distribution $\Pi$ satisfies ${1 \over \nu \, d_1d_2}\le\min_{k, \ell }  \pi_{k \ell}   \le \max_{k,\ell }  \pi_{k \ell }  \leq \f{\mu}{d_1 d_2}$ for some constants $\nu$ and $\mu$. The results show that the max-norm constrained least-squares estimator is indeed rate-optimal in such a setting.

To derive the lower bound, we assume that the sampling distribution $\Pi$ satisfies
\be 
\max_{k,\ell }  \pi_{k \ell }   \leq \f{\mu}{d_1 d_2}    \label{ass2}
\ee
for a positive constant $\mu \geq 1$. Clearly, when $\mu=1$, it amounts to say that the sampling distribution is uniform.

\begin{theorem} \label{Thm3.3}
Suppose that the noise sequence $\{\xi_t\}_{t=1}^n$ are i.i.d. standard normal random variables, the sampling distribution $\Pi$ satisfies the condition \eq{ass2} and the quintuple $(n,d_1, d_2, \alpha,R)$ satisfies
\be  
			\f{48 \alpha^2}{d_1 \vee d_2}			\leq 	R^2    \leq  \f{\sigma^2 (d_1\wedge d_2)d_1 d_2 }{128 \mu n}.   \label{quater}
\ee
Then the minimax $\|\cdot \|_F$-risk is lower bounded as
\be 
	\inf_{\widehat{M}} \sup_{M \in \mathcal{K}(\alpha, R)}   \f{1}{d_1 d_2} \e \| \widehat{M} - M \|_F^2   \geq  \min\left\{  
	\f{\alpha^2}{16},    \f{ \sigma  }{256} R \s{\f{d}{ \mu n}}  \right\}.   \label{minimax-lbd}
\ee
In particular, for a sample size $n \geq  \frac{1}{\alpha^2 \mu}  R^2 d $,
\be 
	\inf_{\widehat{M}} \sup_{M \in \mathcal{K}(\alpha, R)}   \f{1}{d_1 d_2} \e \| \widehat{M} - M \|_F^2   \geq  \f{1}{256}(\alpha \wedge \sigma)  R \s{\f{d}{ \mu n}}.   \label{minimax-lbd2}
\ee
\end{theorem}

Assume that both $\nu$ and $\mu$, respectively appeared in \eq{ass1} and \eq{ass2}, are bounded above by universal constants, then comparing the lower bound \eqref{minimax-lbd2} with the upper bound \eq{mc-ubd3} shows that if the sample size $n >(\frac{R}{\alpha})^2 d$, the optimal rate of convergence is $ R\sqrt{d/n}$; that is,
\be
\label{optimal.rate}
\inf_{\widehat{M}} \sup_{M \in \mathcal{K}(\alpha, R)}   \f{1}{d_1 d_2} \e \| \widehat{M} - M \|_F^2   \asymp   R \s{\f{d_1+d_2}{n}},
\ee
and the max-norm constrained least-squares estimator \eqref{max-est} is rate-optimal. The requirement here on the sample size $n>(\frac{R}{\alpha})^2 (d_1+d_2)$ is weak. If, in addition, $d_1=d_2$, condition \eqref{quater} is reduced to $\alpha^2  d^{-1} \lesssim R^2 \lesssim  \sigma \alpha d$, which is a mild constraint since $R^2$ is of order $\alpha^2 r_0$ in the exact low-rank case where $r_0 = \mbox{rank}(M^*)$.

The proof of Theorem \ref{Thm3.3} uses information-theoretic methods. A key technical tool for the proof is the following lemma which guarantees the existence of a suitably large packing set for $\mathcal{K}(\alpha,R)$ in the Frobenius norm.

\begin{lemma}  \label{cover}
Let $r=(\frac{R}{\alpha})^2 $ and let $\ga \leq 1$ be such that $r \leq \ga^2( d_1\wedge d_2)$ is an integer. Then, there exists a subset $\mathcal{M} \subseteq \mathcal{K}(\alpha, R)$ with cardinality 
$$
	 |\mathcal{M}|  = \left\lfloor \exp\bigg\{ \f{ r (d_1\vee d_2) }{16 \ga^2} \bigg\} \right\rfloor +1
$$
and with the following properties:
\begin{enumerate}
\item[\rm{(i)}] For any $M = ( M_{k \ell} ) \in \mathcal{M}$, \rm{rank}$(M)\leq r/ \ga^2 $ and $M_{k \ell} \in \{ \pm \ga 
\alpha \}$, such that
$$
		\| M \|_{\infty} =  \ga \alpha \leq 1 , \ \  \f{1}{d_1 d_2}\| M \|_F^2 =  \ga^2 \alpha^2.
$$ 
\item[\rm{(ii)}] For any two distinct $M^i, M^j \in \mathcal{M}$, 
\beq 
	\f{1}{d_1 d_2} \| M^i - M^j \|_F^2 > \f{ \ga^2 \alpha^2}{2}.
\eeq
\end{enumerate}
\end{lemma}
The proof of Lemma~\ref{cover} is based on an adaptation of the arguments used to prove Lemma 3 in \cite{DP}, which for self-containment, is given in Section \ref{pf.lemma}.
 
\subsection{Comparison to past work}

We now compare the results established in this section with those known in the literature for matrix completion under uniform or general sampling schemes.

\subsubsection{Approximate/non-exact low-rank recoveries}

It is now well-known that the exact recovery of a low-rank matrix in the noiseless case requires the ``incoherence conditions'' on the target matrix $M^*$ \citep{CR09, CT10, R11, G11}. In this paper, we consider instead a general setting of approximately low-rank matrices, and prove that approximate recovery is still possible without enforcing exact structural assumptions.

Our results are directly comparable to those of \cite{KLT2011} and \cite{NW2012}, in which the trace-norm was used as a proxy to the rank. Taking the latter as an example to illustrate, \cite{NW2012} considered the setup where the sampling distribution is a \textit{product distribution}, i.e. for all $(k,\ell) \in [d_1]\times [d_2]$,
$$
	\pi_{k \ell} = \pi_{k \cdot} \pi_{\cdot \ell},
$$
where $\pi_{k \cdot}$ and $\pi_{\cdot \ell}$ are marginals that satisfy
\be 
	\pi_{k \cdot } \geq \f{1}{\s{\nu}d_1}, \ \  \pi_{ \cdot \ell } \geq \f{1}{\s{\nu}d_2}  \ \ \mbox{ for some } \nu \geq 1. \label{ass1'}
\ee 
Accordingly, define the weighted norms as 
\beq 
	\| M \|_{w(\dagger)} :=  \left\|  \s{W_{{\rm r}} } M \s{W_{{\rm c}}} \right\|_{\dagger}, \ \  \dagger \in \{F, 1 , \infty\},
\eeq 
where $W_{{\rm r}}=d_1 \cdot \mbox{diag}(\pi_{1 \cdot}, \ldots, \pi_{d_1 \cdot})$ and $W_{{\rm c}}=d_2 \cdot \mbox{diag}(\pi_{\cdot 1}, \ldots, \pi_{ \cdot d_2})$

Based on a collection of observations 
$$
	Y_{i_t  j_t} = \varepsilon_t M^*_{i_t  j_t} + \sigma \xi_t, \ \ t=1, \ldots, n,
$$
where $(i_t, j_t)$ are i.i.d. according to $\p \{ (i_t, j_t)=(k,\ell) \} =\pi_{k \ell }$ and $\varepsilon_t \in \{-1 , +1\}$ are i.i.d. random signs, and under the assumption that the unknown matrix $M^*$ satisfies 
\be 
	\| M^* \|_{w(1)} \leq R \s{d_1d_2}, \ \  \| M^* \|_{w(F)} \leq  \s{d_1 d_2} \ \ \mbox{ and } \ \ 
	 \f{\| M^* \|_{w(\infty)}}{\| M^* \|_{w(F)}} \leq \f{\alpha}{\s{d_1 d_2}},  \label{spike.cond}
\ee
\cite{NW2012} proposed the following estimator of $M^*$ based on the trace-norm penalized minimization:
\be 
	\widehat{M}_{{\rm tr}} \in \argmin_{\| M \|_{w(\infty)} \leq  \alpha } \left\{ \f{1}{n}\sum_{t=1}^n (Y_{i_t j_t}- \varepsilon_t  M_{i_t j_t})^2 + \la_n \| M \|_{w(1)} \right\}.	\label{NW-est}
\ee
In the context of low-trace-norm (approximately low-rank) matrix recovery where the true matrix $M^*$ satisfies \eqref{spike.cond}, they proved that for properly chosen $\la_n$ depending on $\sigma$ (see, e.g. Corollary~2 therein), there exist absolute positive constants $c_1$--$c_3$ such that 
\be 
		 \f{1}{d_1 d_2}\| \widehat{M}_{{\rm tr}} - M^* \|_F^2  \leq c_1 \nu \left\{  (\sigma \vee \nu )   \alpha R \s{\f{d \log d}{n}} 
		+  \f{\nu \alpha^2 }{n}  \right\},  \label{NW-r1}
\ee
holds with probability at least $1-c_2\exp(-c_3 \log d)$. 

First, the product distribution assumption can be fairly restrictive in practice and is not valid in many applications. For example, in the case of the Netflix problem, this assumption would imply that conditional on any movie, it will be rated by all users with the same probability. Second, the constraint on $M^*$ highly depends on the true sampling distribution which is really unknown in practice and can only be estimated based on the empirical frequencies, i.e. for any pair $(k,\ell) \in [d_1] \times [d_2]$,
\beq 
	\widehat{\pi}_{k \cdot} = \f{1}{n}\sum_{t=1}^{n}1 \{i_t=k \} , \ \  \widehat{\pi}_{\cdot \ell} = \f{1}{n}\sum_{t=1}^{n}1 \{j_t=\ell \}.
\eeq
Since only a relatively small sample of the entries of $M^*$ is observed, these estimates may not be accurate enough. The max-norm constrained minimization approach, on the other hand, is proved (Theorem \ref{Thm3.1}) to be effective in the presence of non-uniform sampling distributions. The method does not require either a product distribution or the knowledge of  the exact true sampling distribution. From this point of view, the max-norm constrained method indeed yields a more robust approximate recovery guarantee, with respect to the sampling distributions.

We now turn to the special case of uniform sampling. The ``spikeness'' assumption in \cite{NW2012} can actually be reduced to a single constraint on the $\ell_{\infty}$-norm \citep{Klo14}.  Let $\mathbb{B}_{\infty}(\alpha)=\{ M \in \br^{d_1 \times d_2} : \| M \|_{\infty} \leq \alpha\}$ be the $\ell_{\infty}$-norm ball with radius $\alpha$. Define the class of matrices
\be 
	\mathcal{K}_{{\rm tr}}(\alpha, R) := \big\{ M \in \mathbb{B}_{\infty}(\alpha) :  (d_1 d_2)^{-1/2}\| M \|_1 \leq R  \big\}.   \label{tr-set} 
\ee
It can be seen from  \eq{eq2.1} and \eq{eq2.2} that $\{  M \in \mathbb{B}_{\infty}(\alpha) :  \mbox{rank}(M) \leq r \}  \subsetneq  \mathcal{K}(\alpha, \alpha \s{r}) \subsetneq	\mathcal{K}_{{\rm tr}} (\alpha, \alpha\s{r})$. The following results  provide upper bounds on the accuracy of  both the max- and trace-norm regularized estimators under the Frobenius norm.

\begin{corollary} 
Suppose that the noise sequence $\{\xi_t\}_{t=1}^n$ are i.i.d. $N(0,1)$ random variables and the sampling distribution $\Pi$ is uniform on $[d_1]\times [d_2]$. Then the following inequalities hold with probability at least $1-3 d^{-1}$:
\begin{enumerate}
\item[\rm{(i)}] The optimum $\widehat{M}_{\max}$ to the convex program \eq{max-est} satisfies
\be
\sup_{M^* \in \mathcal{K}(\alpha, R)}	\f{1}{d_1 d_2} \| \widehat{M}_{\max} -M^* \|_F^2 \lesssim (  \sigma \vee \alpha ) R  \s{\frac{ d}{n}} 
	+ \alpha^2 \frac{ \log d }{n}.   \label{compare1}
\ee

\item[\rm{(ii)}] The minimum $\widehat{M}_{{\rm tr}}$ to the SDP \eq{NW-est} with all weighted norms replaced by the standard ones and with a {\it properly chosen} $\la_n$ satisfies 
\be
\sup_{M^* \in \mathcal{K}_{{\rm tr}}(\alpha, R)}	\f{1}{d_1 d_2}	 \| \widehat{M}_{{\rm tr}} - M^* \|_F^2  \lesssim   (\sigma \vee \alpha )  R \s{\f{ d \log d }{n}} 
		+ \alpha^2  \frac{ \log d}{n}.  \label{compare2}
\ee
\end{enumerate}
\end{corollary}

The upper bound \eq{compare1} follows immediately from  \eq{mc-ubd} in Theorem \ref{Thm3.1}, and \eq{compare2} is a straightforward extension of Theorem 7 in \cite{Klo14} on exact low-rank matrix recovery to the case of low-trace-norm matrix reconstruction. The proof is essentially the same and thus is omitted.  

\cite{FS2011} analyzed the recovery guarantee for $\widehat{M}_{\max}$ based on an excess risk bound for empirical risk minimization with a smooth loss function recently developed in \cite{SST10}. Specifically, assuming a uniform sampling model with sub-exponential noise and that the target matrix  $M^* \in \mathcal{K}(\alpha,R)$, they proved that with high probability,
\begin{align}
  \frac{1}{d_1 d_2} \| Y - \widehat{M}_{\max} \|_F^2 -   \widehat{\sigma}^2  \lesssim  (\log n)^{3/2} \,   \widehat{\sigma} \sqrt{ \frac{R^2 d }{n}} +  (\log n)^3 \frac{R^2 d}{n}, \label{compare3}
\end{align}
where $Y=M^*+ Z$ with $Z=(\xi_{k \ell})_{1\leq k\leq d_1, 1\leq \ell \leq d_2}$, and $\widehat{\sigma}^2 := \frac{1}{d_1 d_2}\sum_{k=1}^{d_1} \sum_{\ell =1}^{d_2} \xi_{k  \ell}^2$ denotes the average noise level which is concentrated around $\sigma^2$ with high probability.

After a more delicate analysis, our result shows that the additional logarithmic factors in \eq{compare3} purely arise from an artifact of the proof technique and thus can be avoided. Moreover, in view of the lower bounds given in Theorem~\ref{Thm3.3}, we see that the max-norm constrained least square estimator $\widehat{M}_{\max}$ achieves the optimal rate of convergence for recovering approximately low-rank matrices over the parameter space $\mathcal{K}(\alpha, R)$ under the Frobenius norm loss. To our knowledge, the best known rate for the trace-norm regularized estimator given in \eq{compare2} is near-optimal up to logarithmic factors in a minimax sense, over a larger parameter space $\mathcal{K}_{{\rm tr}}(\alpha, R)$.

\subsubsection{Uniform/non-uniform sampling distributions}

We now provide further insight into the rationale behind the phenomenon that the max-norm regularized/constrained method is more robust with respect to the sampling distribution. As before, we focus on the setting with a product sampling distribution $\pi_{k \ell} = \pi_{k \cdot} \pi_{\cdot \ell}$ for $(k,\ell) \in [d_1] \times [d_2]$. 

Motivated by \cite{SS10}, \cite{NW2012} studied the weighted trace-norm penalized estimator $\widehat{M}_{{\rm tr}}$ given at \eqref{NW-est}, where for any matrix $M\in \br^{d_1 \times d_2}$,
\begin{align}
	\| M \|_{w(1)} = \sqrt{d_1 d_2} \left\|  {\rm diag}(\sqrt{\pi_{1\cdot}} , \ldots, \sqrt{\pi_{d_1 \cdot}}) \, M  \, {\rm diag}(\sqrt{\pi_{\cdot 1}} , \ldots, \sqrt{\pi_{ \cdot d_2}})  \right\|_1 . \label{weighted.trace-norm}
\end{align} 
However, the ``true'' form of the sampling distribution is ambiguous and even if it is a product distribution, the marginal probabilities $\pi_{k\cdot}$ and $\pi_{\cdot \ell}$ are typically unknown. Therefore, the weighted trace-norm $\| \cdot \|_{w(1)}$ can not be used in practice.

For the max-norm, a useful equivalent definition is that for any $M \in \br^{d_1\times d_2}$,
\begin{align}
	\| M \|_{\max} = \max_{u\in \br^{d_1}, v\in \br^{d_2}:   \| u \|_2 = \| v\|_2=1 } \left\| {\rm diag}(u) \, M \, {\rm diag}(v) \right\|_1. \nn
\end{align}
See, for example, Theorem~9 in \cite{LSS2008}. As a result, by considering a max-norm penalized estimator that solves 
\begin{align}
\min_{\| M \|_{\infty} \leq  \alpha } \left\{ \f{1}{n}\sum_{t=1}^n (Y_{i_t j_t}- \varepsilon_t  M_{i_t j_t})^2 +   \la_n \| M \|_{\max} \right\}, \nn
\end{align}
all the possible marginal probabilities are taken into account, and therefore the solution is expected to be more robust with respect to the unknown sampling distributions.

Although the sampling distribution is not known exactly in practice, its estimated version is expected to be stable enough as an alternative. According to \cite{FSSS}, given a random sample $S=\{(i_t,j_t)\}_{t=1}^n$, we can estimate $\pi_{k \ell}$ by $\widehat{\pi}_{k \ell } = \widehat{\pi}_{k \cdot} \widehat{\pi}_{\cdot \ell}$ with empirical marginals $\widehat{\pi}_{k \cdot} = n^{-1} \sum_{t=1}^{n}1 \{i_t=k \} $  and $\widehat{\pi}_{\cdot \ell} =  n^{-1}\sum_{t=1}^{n}1 \{j_t=\ell \}$, or by $\widetilde{\pi}_{ij} = \widetilde{\pi}_{k \cdot} \widetilde{\pi}_{\cdot \ell}$ with smoothed empirical marginals
$$
\widetilde{\pi}_{k \cdot} = \f{1}{2} \left( \widehat{\pi}_{k \cdot } +  d_1^{-1} \right), \ \ \widetilde{\pi}_{\cdot \ell }=\f{1}{2}\left(\widehat{\pi}_{\cdot \ell} + d_2^{-1} \right).
$$
The empirically-weighted trace-norm $\| \cdot \|_{ \widehat{w}(1)}$ can be defined in the same spirit as in \eqref{weighted.trace-norm} for the weighted trace-norm, only with $\pi_{k\ell} $ replaced by $\widehat{\pi}_{k \ell}$. Then the unknown matrix can be estimated via penalization on the $\widehat{\pi}$-weighted trace-norm, i.e.
$$
\min_{\| M \|_{\infty} \leq  \alpha } \left\{ \f{1}{n}\sum_{t=1}^n (Y_{i_t j_t}- \varepsilon_t  M_{i_t j_t})^2 +   \la_n \| M \|_{ \widehat{w}(1)}  \right\} .
$$
\cite{FSSS} proved the error bound for the excess risk of the empirically-weighted trace-norm constrained estimator when the loss function is Lipschitz. It is interesting to investigate whether the results similar to those in \cite{NW2012} hold for the empirically-weighted trace-norm constrained and penalized estimators when the quadratic loss function is used.

It is also worth noting that, under condition~\eqref{sub-exp} and when the sampling distribution is nearly uniform in the sense that
\be
	\min_{(k,\ell) \in [d_1] \times [d_2]}\pi_{k \ell} \geq \frac{1}{\nu \, d_1 d_2} \ \ \mbox{ and } \ \   \max \bigg( \sum_{k=1}^{d_1} \pi_{k \ell} , \,
	\sum_{ \ell = 1}^{d_2} \pi_{k \ell }  \bigg) \leq \frac{L}{ \min(d_1, d_2)} \label{near.uniform}
\ee
for some constants $\nu, L \geq 1$, \cite{Klo14} showed that the trace-norm penalized estimator
$$
\widehat{M}_{{\rm tr}} (\lambda) \in \argmin_{\| M \|_{ \infty } \leq  \alpha } \left\{ \f{1}{n}\sum_{t=1}^n (Y_{i_t j_t}- \varepsilon_t  M_{i_t j_t})^2 + \la  \| M \|_{1} \right\}
$$
satisfies 
$$
	\frac{1}{d_1 d_2} \| \widehat{M}_{{\rm tr}} (\lambda)  - M^* \|_F^2 \lesssim (\sigma \vee \alpha)^2 \nu^2 L \frac{r_0 \, d \log d }{n} +  \nu \alpha^2 \sqrt{\frac{\log d}{n}} 
$$
with probability greater than $1-3d^{-1}$, provided that $\| M^* \|_\infty \leq \alpha$ and $\lambda=\lambda_n \asymp \sigma (\frac{L \log d}{nd})^{1/2}$. In the case of Gaussian errors and under condition~\eqref{ass2}, the above rate of convergence is minimax optimal, up to a logarithmic factor, for the class of exact low-rank matrices $\{M \in \br^{d_1\times d_2}: \| M \|_\infty \leq \alpha, \mbox{rank}(M) \leq r_0\}$ \citep{KLT2011}. An interesting and challenging open problem is that in the context of exact low-rank matrix recovery and when the sampling probabilities satisfy \eqref{near.uniform}, whether the optimal recovery guarantee can be achieved using the max-norm constrained method. Also, to the best of our knowledge, there are no theoretical guarantees for exactly recovering a low-rank matrix when the sampling distribution is non-uniform and unspecified.

\section{Computational Algorithms}   \label{implementation.sec}
\setcounter{equation}{0}

Although Theorem \ref{Thm3.1} presents theoretical guarantees that hold uniformly for any global minimizer, it does not provide guidance on how to approximate such a global minimizer using a polynomial-time algorithm. A parallel line of work has studied computationally efficient algorithms for solving problems with the trace-norm constraint or penalization. See \cite{Lin09}, \cite{Maz10} and \cite{Nes07}, among others. Here we restrict our attention to the less-studied max-norm oriented approach. We discuss two different types of algorithms which are particularly designed to solve large scale optimization problems that incorporate the max-norm as a semidefinite relaxation of the rank. The first one is a fast first-order algorithm developed in \cite{Lee2010} based on nonconvex relaxation. The problem of interest to us is the optimization program \eq{max-est} with both the max-norm and the element-wise $\ell_{\infty}$-norm constraints, in which case the algorithm introduced in \cite{Lee2010} can be applied after suitable modifications as described in Section~\ref{alg1}. The second one, on the other hand, is a convex approach proposed by \cite{F2015b} using the alternating direction of multipliers method with guaranteed convergence to the global optimum since it deals with the convex problem \eqref{est2'} directly.

\subsection{A projected gradient method} 
\label{alg1}

Due to \cite{Sre04}, the max-norm of a $d_1 \times d_2$ matrix $M$ can be computed via a semi-definite program:
\[
	\| M \|_{\max}  = \min \, R  \ \ \mbox{ s.t. } \ \  \left(\begin{array}{cc}
	 W_{1} &  M   \\
	M^{\intercal} &  W_{2}   
	\end{array}\right)  \succeq  0,   \ \   \mbox{diag}(W_{1}) \leq R, \ \ \mbox{diag}(W_{2})\leq R .
\]
Correspondingly, we can reformulate \eq{max-est} as the following SDP problem
\begin{align*}
	&\min \, f(M ) \\
	&\mbox{ s.t. } \,  \left(\begin{array}{cc}
	 W_{1} &  M  \\
	M^{\intercal} &  W_{2}   
	\end{array}\right)  \succeq  0,   \ \   \mbox{diag}(W_{1}) \leq R, \ \ \mbox{diag}(W_{2}) \leq R, \  \  \| M \|_{\infty} \leq \alpha ,
\end{align*}
where the objective function $f$ is given by
$$
	f(M) = f(M;Y)= \widehat{\mathcal{L}}_n(M;Y).
$$
This SDP can be solved using standard interior-point methods, though are fairly slow and do not scale to matrices with large dimensions. For large-scale problems, an alternative factorization method based on \eq{eq1.1}, as described below, is preferred \citep{Lee2010}.

We begin by introducing dummy variables $U \in \br^{d_1 \times k}$, $V\in \br^{d_2 \times k}$ for some $1\leq k\leq d_1+d_2$ and let $M=U V^{\intercal}$. If the optimal solution $\widehat{M}_{\max}$ is known to have rank at most $r$, we can take $U \in \br^{d_1 \times (r+1)}$, $V\in \br^{d_2 \times (r+1)}$. In practice, without a known guarantee on the rank of $\widehat{M}_{\max}$, we alternatively truncate the number of columns $k$ to some reasonably high value less that $d_1+d_2$. Then, we rewrite the original problem \eq{max-est} in the factored form as follows:
\beqn 
	\mbox{minimize} && f(UV^{\intercal}) =\f{1}{n}\sum_{t=1}^n (  U_{i_t}^{\intercal} V_{ j_t} - Y_{i_t  j_t} )^2  \ \   \nn \\
	 \mbox{ subject to} &&   \| U \|_{2,\infty}^2 \leq R , \, \| V \|_{2,\infty}^2   \leq  R, \,  \max_{(k,\ell) \in [d_1] \times [d_2]}|U_k^{\intercal} V_\ell | \leq \alpha ,    \label{est2'} 
\eeqn
where $\{(i_1, j_1), \ldots, (i_n,j_n)\} \subseteq ( [d_1] \times [d_2] )^n$ is a training set of row-column indices, $U_i$ and $V_j$ denote the $i$th row of $U$ and the $j$th row of $V$, respectively. This problem, however, is non-convex since it involves a constraint on all product factorizations $UV^{\intercal}$. When the size of the problem $k$ is large enough, \cite{BC06} proved that this reformulated problem has no local minima. To solve this problem fast and efficiently, \cite{Lee2010} suggested the following first-order method. 

Notice that $f(M)=\widehat{\mathcal{L}}_n(M;Y)$ is a smooth function $\br^{d_1\times d_2} \mapsto \br$. The projected gradient descent method generates a sequence of iterates $\{(U^t, V^t), t=0, 1, 2, \ldots \}$ by the recursion: First, define an intermediate iterate
	\[   
	\left[\begin{array}{c}
	\widetilde{U}^{t+1}  \\
	\widetilde{V}^{t+1}
	\end{array}\right]    =  \left[\begin{array}{c}
	U^{t} - \f{\tau}{\s{t}} \cdot  \nabla f(U^t (V^{t })^{\intercal}; Y )V^t \\
	V^{t} - \f{\tau}{\s{t}}  \cdot  \nabla f(U^t (V^{t })^{\intercal}; Y )^{\intercal} U^t
	\end{array}\right]   \ \ \mbox{ for } t= 0, 1, 2, \ldots,
\]
where $\tau>0$ is a stepsize parameter. If $\| \widetilde{U}^{t+1} (\widetilde{V}^{t+1})^{\intercal} \|_{\infty}> \alpha$, we replace 
\[
	\left[\begin{array}{c}
	\widetilde{U}^{t+1}  \\
	\widetilde{V}^{t+1}
	\end{array}\right]   \ \ \mbox{ with } \ \  \f{\s{\alpha}}{\| \widetilde{U}^{t+1} (\widetilde{V}^{t+1})^{\intercal} \|_{\infty}^{1/2}} \left[\begin{array}{c}
	\widetilde{U}^{t+1}  \\
	\widetilde{V}^{t+1}
	\end{array}\right],
\]
otherwise we keep it still. Next, compute updates according to
\[ 
	  \left[\begin{array}{c}
	U^{t+1}  \\
	V^{t+1}
	\end{array}\right]   = \Pi_R \left( \left[\begin{array}{c}
	\widetilde{U}^{t+1}  \\
	\widetilde{V}^{t+1}
	\end{array}\right]  \right), 
\]
where $\Pi_R$ is the Euclidean projection onto $\{(U, V):   \| U \|_{2,\infty}^2 \vee \| V \|_{2,\infty}^2 \leq R\}$. This projection can be computed by re-scaling the rows of the current iterate whose $\ell_2$-norms exceed $R$ so that their norms become exactly $R$, while rows with norms already less than $R$ remain unchanged.

\subsection{An alternating direction method of multipliers based approach}
\label{alg2}

The first-order algorithm described in Section~\ref{alg1} is computationally efficient and fast. 
However, \eqref{est2'} is in principle a non-convex optimization problem and thus the algorithm is only guaranteed to find a stationary point. Recently, an alternating direction method of multipliers (ADMM) based approached was proposed by \cite{F2015b} to solve the convex program \eqref{max-est} efficiently with strong theoretical guarantee. Furthermore, it was shown in \cite{F2015a} that the worst-case rate of convergence of the ADMM method is of order $1/t$, where $t$ denotes the iteration counter. We briefly summarize this ADMM approach here for the sake of readability.

Define the class of matrices
\begin{align}
	\mathcal{P} = \big\{ W \in \mathcal{S}^d : {\rm diag}(W) \geq 0 , \,  \| W_{11} \|_{\infty}\leq R , \, \| W_{22} \|_\infty \leq R , \, \| W_{12} \|_\infty \leq \alpha \big\}, \nn
\end{align}
where $d=d_1 + d_2$, $\mathcal{S}^d$ denotes the class of all symmetric matrices in $\br^{d\times d}$ and for every $W \in \mathcal{S}^d$, we write
\begin{align}
 W =\left(\begin{array}{cc}
	 W_{11} &  W_{12}    \\
	W_{12}^{\intercal} &  W_{22}   
	\end{array}\right) \, \mbox{ with } \, W_{11} \in \br^{d_1 \times d_2}, \ \ W_{22} \in \br^{d_2\times d_2} \, \mbox{ and } \, W_{12} \in \br^{d_1 \times d_2}. \nn
\end{align}
In this notation, the problem \eqref{est2'} can be equivalently formulated as 
\begin{align}
\min_{W, X \in \br^{d\times d}} f(W_{12})  \ \  \mbox{ s.t. } \ \   W \in \mathcal{P} ,  \ \ X \succeq  0  ,  \ \ W - X = 0, \label{prog.2}
\end{align}
where as before, the function $f: \br^{d_1\times d_2} \mapsto \br$ is given by $f(M)=\widehat{L}_n(M;Y)$. As pointed out by \cite{F2015b}, the rationale of reformulating the problem into \eqref{prog.2} is to divide the complexity of the feasible set in \eqref{est2'}, which consists of a positive semidefinite constraint and $\ell_\infty$-norm constraints, into two parts. Then, by using an iterative method, we only need to control the $\ell_\infty$-norm of $W$ and project $X$ into the positive semidefinite cone in each step. The additional constraint $W-X=0$ ensures the feasibility of both $W$ and $X$.

More specifically, consider the augmented Lagrangian function of \eqref{prog.2} that is given by
$$
	F(W, X,  Z ) = f(W_{12}) +2 \langle W - X , Z \rangle + \rho \| W - X \|_F^2
$$
for $ W \in \mathcal{P}$ and $X\in \mathcal{S}^d_+ =\{ S \in \mathcal{S}^d : S \succeq  0 \} $, where $Z$ denotes the dual variable and $\rho>0$ is prespecified. The ADMM is used to solve \eqref{prog.2} iteratively as follows: Initialize $(W^0, X^0, Z^0)$ and $\rho>0$; at the $(t+1)$-th iteration, update $(W,X,Z)$ according to
\begin{align} \label{ADMM.update}
\begin{split}
  X^{t+1} &= \argmin_X   F(W^t, X^t, Z^t) = \Pi_{\mathcal{S}^d_+}( W^t - \rho^{-1} Z^t ), \\
	  W^{t+1} & = \argmin_{W\in \mathcal{P}} \left\{ f(W_{12}) + \rho \| W - X^{t+1} - \rho^{-1} Z^t \|_F^2 \right\},  \\
	  Z^{t+1} & = Z^t + \rho(X^{t+1}-W^{t+1}), \\
\end{split}   
\end{align}
where $\Pi_{\mathcal{S}^d_+}: \br^{d\times d} \mapsto \mathcal{S}^d_+$ is the map that projects a matrix into the semidefinite cone $\mathcal{S}^d_+$. The second step of \eqref{ADMM.update} has an explicit solution given by \citep{F2015b}
\begin{align}
  W^{t+1}_{k\ell} = \begin{cases} 
  	\Pi_{[-\alpha, \alpha]}(\rho^{-1} Y_{k\ell} + W^t_{k\ell} ), \ \ & \mbox{ if } (k,\ell) \in \left(  [d_1] \times [d]\setminus [d_1] \right) \cap S , \\
  		\Pi_{[-\alpha, \alpha]}(  W^t_{k\ell} ), \ \ & \mbox{ if } (k,\ell) \in  \left(  [d_1] \times [d]\setminus [d_1] \right)  \setminus S , \\
  	\Pi_{[-R, R]}(  W^t_{k\ell} ), \ \ & \mbox{ if } (k,\ell) \in  [d_1]\times  [d_1] , \,  k\neq \ell , \\
  	\Pi_{[0, R]}(  W^t_{k\ell} ), \ \ & \mbox{ if } (k,\ell) \in  [d_1]\times [d_1]  ,\, k = \ell,  \\
  		\Pi_{[-R, R]}(  W^t_{k\ell} ), \ \ & \mbox{ if } (k,\ell) \in   [d]\setminus [d_1]  \times  [d]\setminus [d_1] , \, k\neq \ell , \\
  	\Pi_{[0, R]}(  W^t_{k\ell} ), \ \ & \mbox{ if } (k,\ell) \in  [d]\setminus [d_1] \times  [d]\setminus [d_1]   ,\,  k = \ell ,
  \end{cases}  \nn
\end{align}
where $S=\{(i_t,j_t)\}_{t=1}^n$ is the index set of observed entries and $\Pi_{[a,b]}(x) = \min\{ b, \max(a,x) \}$ is the projection function from $\br$ to $[a,b]$. 

\subsection{Implementation}
\setcounter{equation}{0}

Before the max-norm constraint approach can be actually implemented in practice to generate a full matrix by filling in missing entries, additional prior knowledge of the unknown true matrix is needed to avoid deviated results. As before, let $M^* \in \br^{d_1 \times d_2}$ be the true underlying matrix. Suitable upper bounds on the following key quantities are needed in advance:
\be 
		\alpha_0 = \| M^* \|_{\infty}, \ \   R_0 = \| M^* \|_{\max}  \  \   \mbox{ and }  \  \   
		r_0 = \mbox{rank}(M^*).  \lbl{para}
\ee
In order to estimate $R_0$ directly from a missing data matrix, it can be seen from \eq{eq2.2} that $\alpha_0 \s{r_0}$ is a sharp upper bound on $R_0$ and is more amenable to estimation. Fortunately, it is possible to convincingly specify $\alpha_0$ beforehand in many real-life applications. When dealing with the Netflix data, for instance, $\alpha_0$ can be chosen as the highest rating index; in the structure-from-motion problem, $\alpha_0$ depends on the range of the camera field of view, which in most cases is sufficiently large to capture the feature point trajectories. In case where the percentage of missing entries is  low,  the largest magnitude of the observed entries can be used as an alternative for $\alpha_0$.

As for $r_0$, we recommend the rank estimation approach recently developed in \cite{RE11}, which was shown to be effective in computer vision problems. Recall that in the structure-from-motion problem, each column of the data matrix corresponds a trajectory along the frames of a given feature point, and can be regarded as a signal vector with missing coordinates. Due to the rigidity of the moving objects, it was noted in \cite{RE11} that the behavior of observed and missing data is the same and thus they both generate an analogous (frequency) spectral representation. Motivated by this observation, the proposed approach is based on the study of changes in frequency spectra on the initial matrix after missing entries are recovered.

In general, choosing the tuning parameter $R>0$ in \eqref{max-est} adaptively is a difficult problem. In the regression case, it can be done by the Scaled LASSO method \citep{SZ2012}. It is unclear whether a similar approach would work for matrix completion problems. By convexity and strong duality, the optimization program in \eqref{max-est} is equivalent to 
\begin{align}
	\min_{M\in \mathbb{R}^{d_1\times d_2}:   \| M \|_\infty \leq \alpha  } \left\{  \frac{1}{n} \sum_{t=1}^{n} (Y_{i_t j_t} - M_{i_t j_t})^2 + \lambda \| M \|_{\max} \right\}  \label{opt-2}
\end{align}
for a properly chosen $\lambda$. In fact, for any $R>0$ specified in \eqref{max-est}, there exists a $\lambda>0$ such that the solutions to the two problems \eqref{max-est} and \eqref{opt-2} coincide. In practice, we suggest to solve \eqref{opt-2} using the ADMM method described in Section~\ref{alg2} with $\lambda$ obtained via cross-validation, in a way similarly to that for LASSO or the trace-norm penalized $M$-estimator studied in \cite{NW11}.

Next we describe an implementation of the max-norm constrained matrix completion procedure, which incorporates the rank estimation approach in \cite{RE11}. Assume without loss of generality that $\alpha_0$ is known.

\begin{enumerate}
\item[\rm{(1)}] Given the observed partial matrix $M_S$, the initial matrix $M_{{\rm ini}}$ is obtained by adding the average of the corresponding column to the missing entries of $M_S$. Applying the Fast Fourier Transform (FFT) to the columns of $M_{{\rm ini}}$ and taking its modulus, i.e. $F:= |{\rm FFT}(M_{{\rm ini}})|$.

\item[\rm{(2)}] Set an initial rank $r=2$ and an upper bound $r_{\max}$. Clearly, $r_{\max} \leq \min( d_1, d_2)$ and it can be computed automatically by adding a criteria for stopping the iteration.

\item[\rm{(3)}] For the current value of $r$, using the computational algorithms given in Section \ref{implementation.sec} with $R=\alpha_0\s{r}$ to solve the max-norm constraint optimization \eq{max-est}. The resulting estimated full matrix is denoted by $\widehat{M}_r$.

\item[\rm{(4)}] Apply the FFT to $\widehat{M}_r$ as in step 1. Write $F_r = |{\rm FFT}(\widehat{M}_r)|$ and compute the error $e(r)=\|  F- F_r \|_F$.

\item[\rm{(5)}] If $r< r_{\max}$, set $r=r+1$ and go to step 3.

\end{enumerate}

Finally, let 
$$
	r^* = \argmin_{ 2\leq r\leq  r_{\max}} e(r)
$$ 
and the corresponding $\widehat{M}_{r^*}$ is the final estimate of $M^*$. Clearly, the above procedure can be modified by replacing the rank $r$ with the max-norm $R$. A suitable initial value for the max-norm is $R=\alpha_0\s{2}$ and at each iteration, increase $R=R+\de$ with a fixed step size $\de>0$. An upper-bound $R_{\max}$ could be automatically computed by adding some criteria for stopping the iteration.

\section{Discussions}
\label{disc.sec}
\setcounter{equation}{0}

This paper considers the approximate recovery of approximately low-rank matrices, in particular low-max-norm matrices in contrary to low-trace-norm matrices. The max-norm ball with radius 1 is nearly equivalent to the convex hull of rank-1 matrices, and therefore is an alternative convex surrogate for the rank. A max-norm constrained empirical risk minimization method is proposed and its theoretical properties are studied along with computational algorithms. Allowing for \textit{unknown non-uniform sampling} which is an important relaxation of the uniform assumption in practice, it is shown that the method is rate-optimal and can be solved efficiently in polynomial time.

When the underlying matrix has exactly rank $r$, it is known that using the trace-norm based approach leads to a mean square error of order $O\{  r d (\log d)/n\}$ \citep{KAO10, KLT2011, NW2012, Klo14}, where $d=d_1+d_2$. In the ideal uniform sampling model, the trace-norm regularized method is arguably the mostly preferable one as it achieves optimal rate of convergence (up to a logarithmic factor) and is computationally feasible. The sampling scheme considered in this paper is unspecified and is allowed to be highly non-uniform, which brings additional randomness and uncertainty to the recovery problem. Therefore, we are essentially dealing with a much more complex model, and the max-norm constraint is not only introduced as a convex relaxation for low-rankness according to \eqref{eq2.2} but also takes into account the effect of non-uniform sampling.

\section{Proofs}
\label{proof.sec}
\setcounter{equation}{0}

We prove the main results, Theorems \ref{Thm3.1} and \ref{Thm3.3}, in this section. The proofs of a few key technical lemmas including Lemma \ref{cover} are also given.
 
\subsection{Proof of Theorem \ref{Thm3.1}}

For ease of exposition, we write $\widehat{M}=\widehat{M}_{\max}$ as long as there is no ambiguity. To illustrate the main idea, we first consider the case where $\xi_1,\ldots, \xi_n$ are i.i.d. normal random variables and prove that there exists an absolute constant $C$ such that for any $t\in (0,1)$ and a sample size $n$ satisfying $2< n \leq d_1 d_2$,
\be 
	 \| \widehat{M}_{\max} - M^*\|_{\Pi}^2   \leq C\left\{  (\alpha \vee \sigma )R \s{\f{d}{n}} + \alpha^2 \f{ \log (2/t)}{n} \right\}  \label{mc-ubd}
\ee 
holds with probability greater than $1-t - e^{-d} $. The case of sub-exponential noise can be obtained via a straightforward adaptation of the arguments for Gaussian noise. 

To begin with, noting that $\widehat{M}$ is optimal and $M^*$ is feasible for the convex optimization problem \eq{max-est}, we thus have the basic inequality that
$$
	\f{1}{n} \sum_{t=1}^n (Y_{i_t j_t} - \widehat{M}_{i_t j_t} )^2  \leq \f{1}{n} \sum_{t=1}^{n} ( Y_{i_t j_t} - M^*_{i_t j_t} )^2.
$$
This, combined with our model assumption $Y_{i_t j_t}= M^*_{i_t j_t} + \sigma  \xi_t$ yields that
\be
	\f{1}{n}\sum_{t=1}^n \widehat{\De}_{i_t j_t}^2 =\f{1}{n} \sum_{t=1}^n (  \widehat{M}_{i_t j_t} - M^*_{i_t j_t} )^2  \leq \f{2\sigma}{n} \sum_{t=1}^n \xi_t    \widehat{\De}_{i_t j_t},  \label{ineq1}
\ee
where $\widehat{\De}=\widehat{M}-M^* \in \mathcal{K}(2\alpha ,2R)$ is the error matrix. By \eqref{ineq1}, the major challenges in proving Theorem \ref{Thm3.1} consist of two parts, bounding the left-hand side of \eq{ineq1} from below in a uniform sense and the right-hand side of \eq{ineq1} from above.

\medskip
\noi
{\bf Step 1.} (Upper bound). 
Recalling that $\{\xi_t\}_{t=1}^n$ is a sequence of $N(0,1)$ random variables and that $S=\{(i_1, j_1), \ldots, (i_n,j_n)\}$ is drawn i.i.d. according to $\Pi$ on $[d_1] \times [d_2]$, we define
\be
	\widehat{\mathcal{R}}_n(\alpha ,R) := \sup_{M \in \mathcal{K}(\alpha,R)} \left| \f{1}{n} \sum_{t=1}^n \xi_t M_{i_t  j_t} \right|.  \label{sto-err}
\ee
Due to \cite{P89}, we obtain that for any realization of the training set $S$ and for any $\de>0$, with probability at least $1-\de$ over $\bxi=\{\xi_t\}_{t=1}^n$,
\begin{align} 
	 &  \sup_{ M \in \mathcal{K}(\alpha,R)} \left|  \f{1}{n} \sum_{t=1}^n \xi_t M_{i_t j_t} \right|      \nn \\
	& \leq      \e_{\bxi}\left\{  \sup_{M \in \mathcal{K}(\alpha,R) } \left| \f{1}{n}
	\sum_{t=1}^n \xi_t M_{i_t j_t} \right|  \right\}
	+ \pi\s{ \f{\log(1/\de)\sup_{M \in \mathcal{K}(\alpha,R)}\sum_{t=1}^n M_{i_t j_t}^2}{2n^2}} \nn \\
	& \leq      \e_{\bxi}\left\{  \sup_{M \in \mathcal{K}(\alpha,R) } \left| \f{1}{n}\sum_{t=1}^n \xi_t M_{i_t  j_t} \right|  \right\} 
	+ \pi ( \alpha \wedge R ) \s{ \f{\log(1/\de)}{2n}}.   \label{ineq2}
\end{align}
Thus it remains to estimate the following expectation over the class of matrices $\mathcal{K}(\alpha, R)$: 
\beq 
	\mathcal{R}_{n} :=  \e_{\bxi}\left\{  \sup_{M \in \mathcal{K}(\alpha,R) } \left| \f{1}{n}\sum_{t=1}^n \xi_t M_{i_t j_t} \right|  \right\} .
\eeq

As a direct consequence of \eq{eq2.3}, we have
\be 
\mathcal{R}_n \leq K_G \cdot R \cdot  \e_{\bxi} \left(  \max_{M \in \mathcal{M}_{\pm}} \left| \f{1}{n} \sum_{t=1}^n \xi_t M_{i_t  j_t}  \right| \right) ,  \label{sto-err2}
\ee 
where $\mathcal{M}_{\pm}$ contains rank-one sign matrices with cardinality $|\mathcal{M}_{\pm}|=2^{d-1}$.  
For each $M \in \mathcal{M}_{\pm}$, $\sum_{t=1}^n  \xi_t M_{i_t  j_t}$ is a Gaussian random variable with mean zero and variance $n$. Then, the expectation of the Gaussian maximum in \eqref{sto-err2} can be bounded by
\beq 
	2\s{ n \log (|\mathcal{M}_{\pm}| )}  \leq  2\s{ \log 2}\s{n d}.
\eeq
Substituting this into \eqref{sto-err2} gives
\be 
\mathcal{R}_n \leq  2K_G\sqrt{\log 2} \cdot  R\sqrt{nd}  . \nn
\ee 
Since this upper bound holds uniformly over all realizations of $S$, we conclude that with probability at least $1-\de$ over both the random samples $S$ and the noise $\bxi=\{ \xi_t \}_{t=1}^n$, 
\be 
\widehat{\mathcal{R}}_n(\alpha ,R)  \leq  3 \left\{  R \s{\f{d}{n}} + (\alpha \wedge R) \s{\f{\log(1/\de)}{n}}  \right\}.  \label{step1}
\ee

In the case of sub-exponential noise, i.e. $\{ \xi_t \}_{t=1}^n$ satisfies the assumption \eq{sub-exp}, it follows from \eq{eq2.3} that
\beq 
		\widehat{\mathcal{R}}_n(\alpha, R) \leq K_G \cdot R \cdot \sup_{M \in \mathcal{M}_{\pm} } \left| \f{1}{n} \sum_{t=1}^n \xi_t M_{i_t  j_t} \right| \ \ \mbox{ with } \ \  		 |\mathcal{M}_{\pm}| = 2^{d-1}.
\eeq
For any realization of the training set $S=\{(i_1, j_1), \ldots, (i_n, j_n)\}$ and for any $M \in \mathcal{M}_{\pm}$ fixed, it follows from a Bernstein-type inequality for sub-exponential random variables \citep{Ver12} that
\beq 
	\p \left(  \left| \f{1}{n} \sum_{t=1}^n \xi_t M_{i_t  j_t} \right|  \geq t \right)  \leq 2 \exp\left\{  -c \cdot \min \left(  \f{n t^2}{K^2} , \f{nt}{K} \right)  \right\},
\eeq
where $c>0$ is an absolute constant. By the union bound, it can be easily verified that for a sample size $n\geq d$,
\be
	\widehat{\mathcal{R}}_n(\alpha, R)  \leq C K R  \s{\f{d}{n}}   \label{step1-gen}
\ee
holds with probability at least $1-e^{-d}$ for some absolute constant $C>0$.

\medskip

\noi
{\bf Step 2.} (Lower bound).
For the given sampling distribution $\Pi$, note that
$$
\| M \|_{\Pi}^2  = \sum_{k, \ell } \pi_{k \ell }M^2_{k \ell} =  \frac{1}{|S|} \e_{S \sim \Pi} \| M_S \|_2^2 ,
$$
where $M_S=(M_{i_1  j_1}, \ldots, M_{i_n  j_n})^{\intercal} \in \br^n$ for any training set $S=\{(i_t, j_t)\}_{t=1}^n$ of size $n$. For $\beta \geq 1$ and $\de>0$, consider the following subset
\beq 
	\mathcal{C}(\beta, \de) := \big\{ M \in \mathcal{K}(1, \beta) : \| M \|_{\Pi}^2 \geq \de \big\}.
\eeq
Here, $\de$ can be regarded as a tolerance parameter. The goal is to show that there exists some function $f_{\beta}$ such that with high probability, the following inequality 
\be
	\f{1}{n}\| M_S \|_2^2 \geq \f{1}{2}\| M \|_{\Pi}^2 - f_{\beta}(n,d_1, d_2)    \label{RSC}
\ee
holds uniformly over $M \in \mathcal{C}(\beta,\de)$.

\medskip
\noi
{\bf Proof of \eq{RSC}.} Instead, we will prove a stronger result that with exponentially high probability,
\beq 
 \left| \f{1}{n} \| M_S \|_2^2 - \| M \|_{\Pi}^2 \right| \leq \f{1}{2}\| M \|_{\Pi}^2 + f_{\beta}(n,d_1, d_2)
\eeq
holds for all $M \in \mathcal{C}(\beta, \de)$, based on a straightforward adaptation of the peeling argument used in \cite{NW2012}. Taking $\varrho=\f{3}{2}$, define a sequence of subsets
\beq 
	   \mathcal{C}_{\ell}(\beta, \de) := \big\{ M \in \mathcal{C}(\beta, \de):   \varrho^{\ell -1} \de  \leq \| M \|_{\Pi}^2 \leq \varrho^{\ell} \de  \big\}
\eeq
for $\ell=1,2, \ldots,$ and for any radius $D>0$, set
\be 
	 \mathcal{B}(D) := \left\{ M \in \mathcal{C}( \beta, \de):   \| M \|_{\Pi}^2  \leq D  \right\}.  \label{setBD}
\ee
In fact, if there exists some $M \in \mathcal{C}(\beta, \de)$ satisfying
\beq 
	 \left|\f{1}{n}\|M_S \|_2^2- \| M \|_{\Pi}^2  \right| > \f{1}{2}\| M \|_{\Pi}^2 + f_{\beta}(n,d_1, d_2),
\eeq
then there corresponds an $\ell \geq 1$ such that, $M \in \mathcal{C}_{\ell}(\beta,\de) \subseteq \mathcal{B}(\varrho^{\ell} \de)$ and
\beq 
	 \left| \f{1}{n}\|M_S \|_2^2- \| M \|_{\Pi}^2  \right|  > \f{1}{3} \varrho^{\ell} \de + f_{\beta}(n,d_1, d_2).
\eeq
Therefore, the main task is to show that the latter event occurs with small probability. To this end, define the maximum deviation for each $S\subseteq ([d_1]\times [d_2])^n$ that
\be
	\De_{D}(S)  = \sup_{M \in \mathcal{B}(D) }  \left| \frac{1}{n}\| M_S \|_2^2- \| M \|_{\Pi}^2 \right| .  \label{max-dev}
\ee
The following lemma shows that $n^{-1}\| M_S \|_2^2$ does not deviate far from its expectation {\it uniformly} for all $M \in \mathcal{B}(D)$.

\medskip
\begin{lemma}[Concentration]  \label{badevents}
There exists a universal positive constant $C_1$ such that, for any $D>0$, 
\be
 \p\left\{ \De_{D}(S) > \f{D}{3} +C_1 \beta \s{\f{d}{n}} \right\} \leq e^{-nD/26}.  \label{prob-be}
\ee
\end{lemma}

\medskip

In view of the above lemma, we take $f_{\beta }(n,d_1, d_2) =C_1 \beta \sqrt{d/n}$ and consider the following sequence of events
\beq 
	\mathcal{E}_{\ell} = \big\{  \De_{\varrho^{\ell} \de}(S) >  \tfrac{1}{3}\varrho^{\ell} \de  + f_{\beta}(n,d_1, d_2) \big\}  \ \ \mbox{ for }  \ell= 1, 2, \ldots .
\eeq
Because $\mathcal{C}( \beta , \de)= \cup_{ \ell \geq 1} \mathcal{C}_{\ell}(\beta ,\de)$, using the union bound we have
\begin{align}
	&   \p \left\{ \exists M \in \mathcal{C}( \beta ,\de), \mbox{ s.t. }    \left| \f{1}{n}\| M_S \|_2^2- \| M \|_{\Pi}^2  \right| > \f{1}{2}\| M \|_{\Pi}^2 +  f_{\beta }(n,d_1, d_2)   \right\}  \nn \\
	&  \leq  \sum_{\ell =1}^\infty \p \left\{ \exists M \in \mathcal{C}_{\ell}(\beta ,\de), \mbox{ s.t. }  \left| \f{1}{n}\| M_S \|_2^2- \| M \|_{\Pi}^2  \right| > \f{1}{2}\| M \|_{\Pi}^2 +  f_{\beta }(n,d_1, d_2)   \right\}  \nn \\
	&    \leq  \sum_{\ell =1}^{\infty} P \left( \mathcal{E}_{\ell}^c \right)  \nn \\
	&  \leq \sum_{\ell =1}^{\infty} \exp(-n \varrho^{\ell} \de/26)  \nn \\
	&  \leq \sum_{\ell =1}^{\infty} \exp\{ -\log(\varrho) \ell n \de /26 \} \leq \f{\exp(-c_0 n  \de)}{1-\exp(-c_0 n  \de)}
\end{align}
with $c_0 = \log(3/2)/26$, where we used the elementary inequality that 
$$
	\varrho^{\ell}  = \exp\{\ell \log(\varrho)\} \geq  \ell  \log(\varrho).
$$ 
Consequently, for a sample size $n \leq d_1 d_2$ satisfying $\exp(-c_0   n \de)\leq \f{1}{2}$, or equivalently, $n> (c_0   \de)^{-1} \log 2 $, we obtain that with probability greater than $1-2\exp(-c_0  n  \de)$,
\be
	\f{1}{n} \| M_S \|_2^2 \geq \f{1}{2} \| M \|_{\Pi}^2 - C_1 \beta \s{\f{d}{n}}    \label{step2}
\ee
holds for all $M \in \mathcal{C}(\beta , \de) $.

\medskip
\noi
{\bf Step 3.} Now we combine the results in {\it Step 1} and {\it Step 2} to finish the proof. On one hand, it follows from \eq{step1} that for a sample size $2<n\leq d_1d_2$,
\be 
	\f{1}{n}\sum_{t=1}^n \xi_t  \widehat{\De}(i_t, j_t)  \leq   \widehat{\mathcal{R}}_n(2\alpha, 2R) \leq 
	12 R  \s{\f{d}{n}} \nn
\ee 
holds with probability at least $1-e^{-d}$. On the other hand, set $\widetilde{\De}=\widehat{\De}/(2\alpha)$ such that $\|\widetilde{\De} \|_{\infty} \leq 1$ and $\| \widetilde{\De}\|_{\max}\leq R/\alpha := \beta$, or equivalently, $\widetilde{\De}\in \mathcal{K}(1,\beta)$. For any $0<t<1$, applying \eq{step2} with $\de=\f{\log(2/t)}{c_0 n}$ implies that for a sample size $n$ with $2< n \leq d_1 d_2$,
\beq
	\| \widetilde{\De} \|_{\Pi}^2  \leq \max\left\{ \f{\log(2/t)}{c_0 n} ,    \f{2}{n}\| \widetilde{\De}_S \|_2^2 + 2\beta C_1\s{\f{d}{n}}  \right\}
\eeq
holds with probability at least $1-t$. The last two displays, joint with the basic inequality \eq{ineq1} lead to the final conclusion \eq{mc-ubd} after a simple rescaling. Similarly, using the upper bound \eq{step1-gen}, instead of \eq{step1}, together with the lower bound \eq{step2} proves \eq{mc-ubd-gen} in the case of sub-exponential noise.  \qed  \\

\subsubsection{Proof of Lemma \ref{badevents}}

Here, we prove the concentration inequality given in Lemma \ref{badevents}. The argument is based on some basic techniques of probability in Banach spaces, including symmetrization, contraction inequality and Bousquet's version of Talagrand concentration inequality as well as the upper bound \eq{eq2.4} on the empirical Rademacher complexity of the max-norm ball.

Regarding the matrix $M \in \br^{d_1 \times d_2}$ as a function: $[d_1] \times [d_2] \mapsto \br$, i.e. $M(k,\ell)=M_{k \ell}$, we are interested in the following empirical process indexed by $\mathcal{B}(D)$:
\beq 
	\De_{D}(S) = \sup_{f_M: M \in \mathcal{B}(D)} \left| \f{1}{n} \sum_{t=1}^n f_M(i_t, j_t) - \e \{ f_M(i_t, j_t) \}  \right| \ \ \mbox{ with } \ \  	f_M(\cdot ) =  \{ M( \cdot ) \}^2.
\eeq
Recall that $|M_{k \ell }| \leq \| M \|_{\infty} \leq  1$ for all pairs $(k, \ell )$, we have
$$
	   \sup_{M \in \mathcal{B}(D)} \var \{ f_M(i_1, j_1) \} \leq \sup_{M \in \mathcal{B}(D)} \| M \|_{\infty}^2 \| M \|_{\Pi}^2 \leq  D.
$$
We first bound $\e_{S \sim \Pi} \{ \De_{D}(S) \}$, and then show that $\De_{D}(S)$ is concentrated around its expectation. A standard symmetrization argument \cite{LT91} yields
\beq 
	\e_{S \sim \Pi} \{ \De_{D}(S) \} \leq 2 \e_{S\sim \Pi}\left[ \e_{ \varepsilon} \left\{  \sup_{M \in \mathcal{B}(D) }  \left| \f{1}{n}\sn \varepsilon_i 	M_{i_t j_t}^2 \right|   \right\}  \right],
\eeq
where $\{\varepsilon_i\}_{i=1}^n$ is an i.i.d. Rademacher sequence, independent of $S$. Given an index set $S=\{(i_1, j_1), \ldots, (i_n, j_n)\}$, since $|M_{i_t  j_t}| \leq 1 $, using Ledoux-Talagrand contraction inequality \citep{LT91} implies that for $d=d_1+d_2$,
\begin{align}
 & \e_{ \varepsilon} \left\{  \sup_{M \in \mathcal{B}(D) }  \left| \f{1}{n}\sn \varepsilon_i
	M_{i_t j_t}^2 \right|  \right\}    \leq   4  \e_{\varepsilon} \left\{  \sup_{M \in \mathcal{B}(D) } \left|  \f{1}{n}
\sum_{t=1}^n \varepsilon_t M_{i_t  j_t} \right|   \right\}  \nn \\
& \leq  4  \e_{\varepsilon} \left(  \sup_{\| M \|_{\max} \leq \beta} \left|  \f{1}{n} \sum_{t=1}^n \varepsilon_t M_{i_t j_t}\right|   \right) \leq   48 \beta \s{\f{d}{n}}, \nn
\end{align}
where we used inequality \eq{eq2.4} in the last step. Since the ``worst-case'' Rademacher complexity is uniformly bounded, we have
\be 
	\e_{S \sim \Pi} \{ \De_D(S) \}  \leq 96 \beta \s{\f{d}{n}}.  \label{mean-bd}
\ee

Next, applying Bousquet's version of Talagrand's concentration inequality for empirical processes indexed by bounded functions \citep{B2003} yields that for every $t>0$,
\begin{align}
	\De_{D}(S) &   \leq   \e_{S \sim \Pi} \{ \De_{D}(S) \} + \s{\frac{2tD}{n} +  \e_{S \sim \Pi} \{ \De_{D}(S) \} \frac{4t}{n} }+ \frac{ t}{3n}   \nn \\
	& \leq   \e_{S \sim \Pi} \{ \De_{D}(S) \}  +   2 \sqrt{ \e_{S \sim \Pi} \{ \De_{D}(S) \} \frac{t}{n}}    + \sqrt{\frac{2tD}{n}} +  \frac{ t}{3n}  \nn \\
	& \leq 2\e_{S \sim \Pi} \{ \De_{D}(S) \} +  \sqrt{\frac{2tD}{n}} +  \frac{ 4t}{3n}   \nn
\end{align}
with probability at least $1-  e^{-t}$. The conclusion \eq{prob-be} thus follows by taking $t=n D /26$.   \qed 

\subsection{Proof of Theorem \ref{Thm3.2}}

The proof is based on a general result in \cite{SST10} on excess risk bounds for learning with a smooth loss. Recall that the noisy response is of the form $Y_{i_t j_t} = M^*_{i_t j_t} + \xi_t$ for $t=1, 2, \ldots$, where the location $(i_t, j_t)$ of the entry is drawn from $[d_1]\times [d_2]$ according to $\Pi$ and the noise $\xi_t$ on the entry is drawn independently each time. For every $d_1 \times d_2$ matrix $M$, define the quadratic loss function
\begin{align*}
 \mathcal{L}(M) & =  \e_{(i_t, j_t) \sim \Pi \atop \xi_t \sim N(0,1)} (M- Y )_{i_t j_t}^2 \\
&  =\e_{(i_t, j_t) \sim \Pi \atop \xi_t \sim N(0,1)}  \{ (M^* - M )_{i_t j_t} + \xi_t \}^2   = \| M - M^* \|^2_{\Pi} + \sigma^2 ,
\end{align*}
and its empirical counterpart $\widehat{\mathcal{L}}(M) = \frac{1}{n} \sum_{t=1}^n (M_{i_t j_t}- Y_{i_t j_t})^2 + \sigma^2 $ for a given i.i.d. sample $\{(i_t, j_t), Y_{i_t j_t} = M^*_{i_t j_t}+ \xi_t\}_{t=1}^n$. In this notation, our estimator $\widehat{M}_{\max}$ can be written as $\widehat{M}_{\max} = \argmin_{M\in \mathcal{K}(\alpha, R)} \widehat{\mathcal{L}}(M)$.

In view of Definition~\ref{Def2.1}, define the worst-case Rademacher complexity as
\begin{align}
	 R_n(\mathcal{K}) & = \sup_{  \{ (i_t, j_t) \}_{t=1}^n \in ([d_1]\times [d_2])^{n} } \e_{\beps} \left\{ \sup_{M\in \mathcal{K}} \frac{1}{n} \left| \sn \varepsilon_i M(i_t, j_t ) \right| \right\} \nn \\
	 &  =  \sup_{  \{ (i_t, j_t) \}_{t=1}^n \in ([d_1]\times [d_2])^{n} } \e_{\beps} \left( \sup_{M\in \mathcal{K}} \frac{1}{n} \left| \sn \varepsilon_i M_{ i_t   j_t } \right| \right), \nn
\end{align}
where $\mathcal{K} = \mathcal{K}(\alpha, R)$.

For any $B>0$, let $\mathcal{E}_B$ be the event that $ \max_{1\leq t\leq n}|\xi_t |\leq B$ holds. On $\mathcal{E}_B$, applying Theorem~1 in \cite{SST10} by taking $H=2$ and $b=5\alpha^2 + 4\alpha \sigma B$ that, for any $0<\delta<1$,
\begin{align}
	& \mathcal{L}(\widehat{M}_{\max} )  -  \min_{M\in \mathcal{K}(\alpha, R)} {\mathcal{L}}(M)  \nn \\
	&  \leq  C_1 \bigg[  \sqrt{ \min_{M\in \mathcal{K}(\alpha, R)} {\mathcal{L}}(M)\bigg\{  (\log n)^3 R^2_n(\mathcal{K}) + \frac{B\log(1/\delta)}{n} \bigg\} }  \nn \\
	& \qquad \qquad +  (\log n)^3 R^2_n(\mathcal{K}) + \frac{B\log(1/\delta)}{n} \bigg] \nn
\end{align}
holds with probability at least $1-\delta$ over a random sample $\{(i_t, j_t)\}_{t=1}^n$ of size $n$, where $C_1>0$ is an absolute constant. By \eqref{eq2.4}, the worst-case Rademacher complexity $R_n(\mathcal{K})$ is bounded by $6R\sqrt{d/n}$. Moreover, note that $\min_{M\in \mathcal{K}(\alpha, R)} {\mathcal{L}}(M) =  \mathcal{L}(M^*) = \sigma^2$ and
\begin{align}
	 \mathcal{L}(\widehat{M}_{\max} )  = \| \widehat{M}_{\max} - M^* \|_{\Pi}^2 + \sigma^2. \nn
\end{align}

Putting the above calculations together, we obtain that on the event $\mathcal{E}_B$,
\begin{align}
	&   \| \widehat{M}_{\max} - M^* \|_{\Pi}^2  \nn \\
	& \leq     C_2 \,\sigma \left[  \sqrt{  \left\{  (\log n)^3 \frac{R^2 d}{n} + \frac{B\log(1/\delta)}{n} \right\} }  +  (\log n)^3 \frac{R^2 d}{n}  + \frac{B\log(1/\delta)}{n} \right]    \label{excess.risk.bound.1}
\end{align}
holds with probability at least $1-\delta$.

Finally, it follows from Borell's inequality that for every $t>0$,
\begin{align}
	 \P \left\{ \max_{1\leq t\leq n} |\xi_t| \geq  \e \left( \max_{1\leq t\leq n} |\xi_t| \right) +  t \right\} \leq e^{-t^2/2}. \nn
\end{align}
A standard result on Gaussian maximum gives $\e ( \max_{1\leq t\leq n} |\xi_t|  )  \leq 2\sqrt{\log n}$. Together with the last display, this implies that with probability at least $1-\delta$,
\be
	 \max_{1\leq t\leq n} |\xi_t| \leq 2\sqrt{\log n} + \sqrt{2\log(1/\delta)}. \label{noise.concentration}
\ee
In particular, taking $\delta= n^{-1}$ in both \eqref{excess.risk.bound.1} and \eqref{noise.concentration} proves \eqref{error.bound.2}. \qed

\subsection{Proof of Theorem \ref{Thm3.3}}

By construction in Lemma \ref{cover}, setting $\de=\ga  \alpha \s{d_1 d_2/2}$ we see that $\mathcal{M}$ is a $\de$-packing set of $\mathcal{K}(\alpha, R)$ in the Frobenius norm. Next, a standard argument \citep{YB99, Yu97} yields a lower bound on the $\| \cdot \|_F$-risk in terms of the error in a multi-way hypothesis testing problem. More specifically,
\beq 
	\inf_{\widehat{M}} \max_{M \in \mathcal{K}(\alpha, R) } \e \| \widehat{M} - M \|_F^2 \geq \f{\de^2}{4} \min_{\widetilde{M}} \p(\widetilde{M} \neq M^*),
\eeq
where the random variable $M^* \in \br^{d_1 \times d_2}$ is uniformly distributed over the packing set $\mathcal{M}$. Conditional on $S=\{(i_1, j_1), \ldots, (i_n, j_n)\}$, a variant of Fano's inequality \citep{fano} leads to the lower bound
\be 
	\p(\widetilde{M} \neq M^* | S)  \geq 1- \f{{|\mathcal{M}| \choose 2}^{-1}\sum_{i \neq j} K(M^i \| M^j) +  \log 2}{\log |\mathcal{M}|},  \lbl{fi}
\ee 
where $K(M^i \| M^i)$ denotes the Kullback-Leibler divergence between distributions $(Y_S|  M^i)$ and $(Y_S|  M^j)$. For the observation model \eq{mc-md} with i.i.d. Gaussian noise, we have
\beq 
K(M^i \| M^j) = \f{1}{2\sigma^2}\sum_{t=1}^n (M^i - M^j)^2_{i_t j_t}
\eeq
and 
\be
	\e_{S \sim \Pi} \{ K(M^i \| M^j) \}  = \f{n}{2\sigma^2} \| M^i -M^j \|_{\Pi}^2,  \label{exp-KL}
\ee
where $\| \cdot \|_{\Pi}$ is the weighted Frobenius norm as in \eqref{weighted.Fro.norm}. For any two distinct $M^i, M^j \in \mathcal{M}$, $\| M^i - M^j \|_F^2 \leq 4d_1 d_2 \ga^2$, which together with \eq{fi}, \eq{exp-KL} and the assumption $\max_{k,\ell} \pi_{k \ell} \leq \f{\mu}{d_1 d_2}$ implies that
\begin{align}
	& \p(\widetilde{M} \neq M^* ) \nn \\
	 &  \geq  1- \f{{|\mathcal{M}| \choose 2}^{-1}\sum_{i \neq j} \e_{S \sim \Pi}\{ K(M^i \| M^j) \} +  \log 2}{\log |\mathcal{M}|}     \nn \\
	& \geq    1- \f{\f{32 \mu \ga^4 \alpha^2 n}{\sigma^2} + 12 \ga^2}{ r (d_1 \vee d_2)} \geq 1-\f{32 \mu \ga^4 \alpha^2 n}{\sigma^2  r (d_1 \vee d_2)}- \f{12}{ r (d_1 \vee d_2)} \geq \f{1}{2},
\end{align}
provided that $ r (d_1 \vee d_2) \geq 48$ and $\ga^4 \leq \f{\sigma^2 }{128  \alpha^2 } \f{r (d_1 \vee d_2)}{\mu n}$. If $\f{\sigma^2 }{128  \alpha^2 } \f{r (d_1 \vee d_2) }{\mu n} >1$, we choose $\ga=1$ so that 
\beq 
	\inf_{\widehat{M}} \max_{M \in \mathcal{K}(\alpha, r)}  \f{1}{d_1d_2}\e \| \widehat{M} - M \|_F^2 
	\geq \f{\alpha^2}{16}.
\eeq
Otherwise, as long as the parameters  $(n, d_1, d_2, \alpha, R)$ satisfy \eq{quater}, taking
$$
	\ga^2=\f{\sigma }{8\s{2} \,\alpha}\s{\f{r (d_1 \vee d_2)}{ \mu n }}
$$ 
yields
\beq 
	\inf_{\widehat{M}} \max_{M \in \mathbb{B}_{\max}(R)}  \f{1}{d_1d_2}\e \| \widehat{M} - M \|_F^2 
	\geq \f{ \sigma \alpha}{128\s{2}} \s{\f{ r(d_1 \vee d_2)}{\mu n}} \geq \f{\sigma R}{256}\s{\f{d}{\mu n}},
\eeq
as desired. \qed    \\

\subsection{Proof of Lemma \ref{cover}}   
\lbl{pf.lemma}

We proceed via a probabilistic method. Assume without loss of generality that $d_2 \geq d_1$. Let $N=\exp(\f{r d_2}{16 \ga^2})$, $B=\f{r}{\ga^2}$, and for each $i=1, \ldots, N$, we draw a random matrix $M^i \in \br^{d_1 \times d_2}$ as follows: The matrix $M^i$ consists of i.i.d. blocks of dimensions $B \times d_2$, stacked from top to bottom, with the entries of the first block being i.i.d. symmetric random variables taking values $\pm \alpha \ga$, such that 
\beq 
	M^i_{k \ell} := M^i_{k' \ell }, \ \  k' = k(\mbox{mod } B) +1.
\eeq 
Next, we show that above random procedure succeeds in generating a set having all desired properties, with non-zero probability. For $1\leq i\leq N$, it is easy to see that 
$$
	\| M^i \|_{\infty} = \alpha \ga \leq \alpha, \ \  \f{1}{d_1 d_2}\| M^i \|_F^2 =  \alpha^2 \ga^2
$$ 
and because rank$(M^i)\leq B$,
\beq 
	\| M^i \|_{\max} \leq  \s{B} \, \| M^i \|_{\infty} = \s{\f{r}{\ga^2}} \,\alpha \ga = \alpha \s{r} =R.
\eeq
Consequently, $M^i \in \mathcal{K}(\alpha, R)$ and it remains to show that the set $\{M^i\}_{i=1}^N$ satisfies property \rm{(ii)}. In fact, for any $1\leq i \neq j\leq N$,
\begin{align*}
 \| M^i - M^j \|_F^2  & = \sum_{k, \ell} (M^i_{k\ell}-M^j_{k\ell})^2  \\
 & \geq   \left \lfloor \f{d_1}{B} \right\rfloor  \sum_{k=1}^B \sum_{\ell =1}^{d_2} (M^i_{k\ell }-M^j_{k\ell })^2  =  4\alpha^2 \ga^2  \left\lfloor \f{d_1}{B} \right\rfloor   \sum_{k=1}^B \sum_{\ell =1}^{d_2} \de_{kl},
\end{align*}
where $\de_{kl}$ are independent $0/1$ Bernoulli random variables with mean $1/2$. Using Hoeffding's inequality gives
\beq 
	\p \left(   \sum_{k=1}^B \sum_{\ell =1}^{d_2} \de_{k \ell}  \geq \f{ Bd_2}{4} \right)  \leq \exp(-  B d_2 /8).
\eeq
Because there are less than $N^2/2 $ such index pairs in total, the above inequality, together with the union bound implies that with probability at least $1-\f{N^2}{2}\exp(- Bd_2 /8) \geq 1/2 $,
\beq 
	\| M^i - M^j \|_F^2 > \alpha^2 \ga^2  \left\lfloor \f{d_1}{B}  \right\rfloor   B d_2  \geq \f{\alpha^2 \ga^2 d_1 d_2}{2} 
\eeq
holds for all $i\neq j$. This completes the proof of Lemma \ref{cover}.  \qed

\section*{Acknowledgements}
We thank the editors and an anonymous referee for their careful reviews and constructive comments.

\end{document}